\documentclass[12pt]{article}

\usepackage{newtxtext,newtxmath}
\usepackage{placeins}
\usepackage[hidelinks, bookmarks=true, bookmarksopen=true]{hyperref}
\usepackage{graphicx}

\usepackage[letterpaper,margin=1in]{geometry}

\renewenvironment{abstract}
	{\quotation}
	{\endquotation}

\date{}

\makeatletter
\renewcommand{\fnum@figure}{\textbf{Figure \thefigure}}
\renewcommand{\fnum@table}{\textbf{Table \thetable}}
\makeatother

\usepackage{scicite}

\usepackage{url}

\usepackage{amsmath}

\newcommand{\indep}{\mathrel{\perp\!\!\!\perp}}

\def\scititle{
\vspace{-2.1cm}
	RAPTOR: A Foundation Policy for Quadrotor Control
}
\title{\bfseries \boldmath \scititle}

\author{
	Jonas~Eschmann$^{1\ast}$,
	Dario~Albani$^{2}$,
	Giuseppe~Loianno$^{1}$\and
	\small$^{1}$EECS, UC Berkeley, USA.\and
	\small$^{2}$ARRC, Technology Innovation Institute, UAE.\and
	\small$^\ast$Corresponding author. Email: jonas.eschmann@berkeley.edu
}
\begin{document}

% Insert the title and author list
\vspace{-10.1cm}
\maketitle

% Abstract, in bold
% There are strict length limits, and not all formats have abstracts.
% Consult the journal instructions to authors for details.
% Do not cite any references in the abstract.
\vspace{-10mm}

\begin{center}
\textbf{Project Page}: \url{https://raptor.rl.tools}

\textbf{Video}: \url{https://youtu.be/hVzdWRFTX3k}
\end{center}

% \begin{quote}
% \centering
% \textbf{Summary:}
% RAPTOR is a tiny, end-to-end, neural control policy with 2084 parameters that adapts zero-shot to 10 unseen real quadrotors via in-context learning.
% \end{quote}

\begin{abstract} \bfseries \boldmath
\vspace{1mm}
Humans are remarkably data-efficient when adapting to new unseen conditions, like driving a new car. In contrast, modern robotic control systems, like neural network policies trained using Reinforcement Learning (RL), are highly specialized for single environments. Because of this overfitting, they are known to break down even under small differences like the Simulation-to-Reality (Sim2Real) gap and require system identification and retraining for even minimal changes to the system. In this work, we present RAPTOR, a method for training a highly adaptive foundation policy for quadrotor control. Our method enables training a single, end-to-end neural-network policy to control a wide variety of quadrotors. We test $10$ different real quadrotors from $32\ \mathrm{g}$ to $2.4\ \mathrm{kg}$ that also differ in motor type (brushed vs. brushless), frame type (soft vs. rigid), propeller type (2/3/4-blade), and flight controller (PX4/Betaflight/Crazyflie/M5StampFly). We find that a tiny, three-layer policy with only $2084$ parameters is sufficient for zero-shot adaptation to a wide variety of platforms. The adaptation through in-context learning is made possible by using a recurrence in the hidden layer. The policy is trained through our proposed Meta-Imitation Learning algorithm, where we sample $1000$ quadrotors and train a teacher policy for each of them using RL. Subsequently, the $1000$ teachers are distilled into a single, adaptive student policy. We find that within milliseconds, the resulting foundation policy adapts zero-shot to unseen quadrotors. We extensively test the capabilities of the foundation policy under numerous conditions (trajectory tracking, indoor/outdoor, wind disturbance, poking, different propellers).
\vspace{-2.5mm}
\end{abstract}

\begin{figure}
	\centering
	\includegraphics[width=1.0\textwidth]{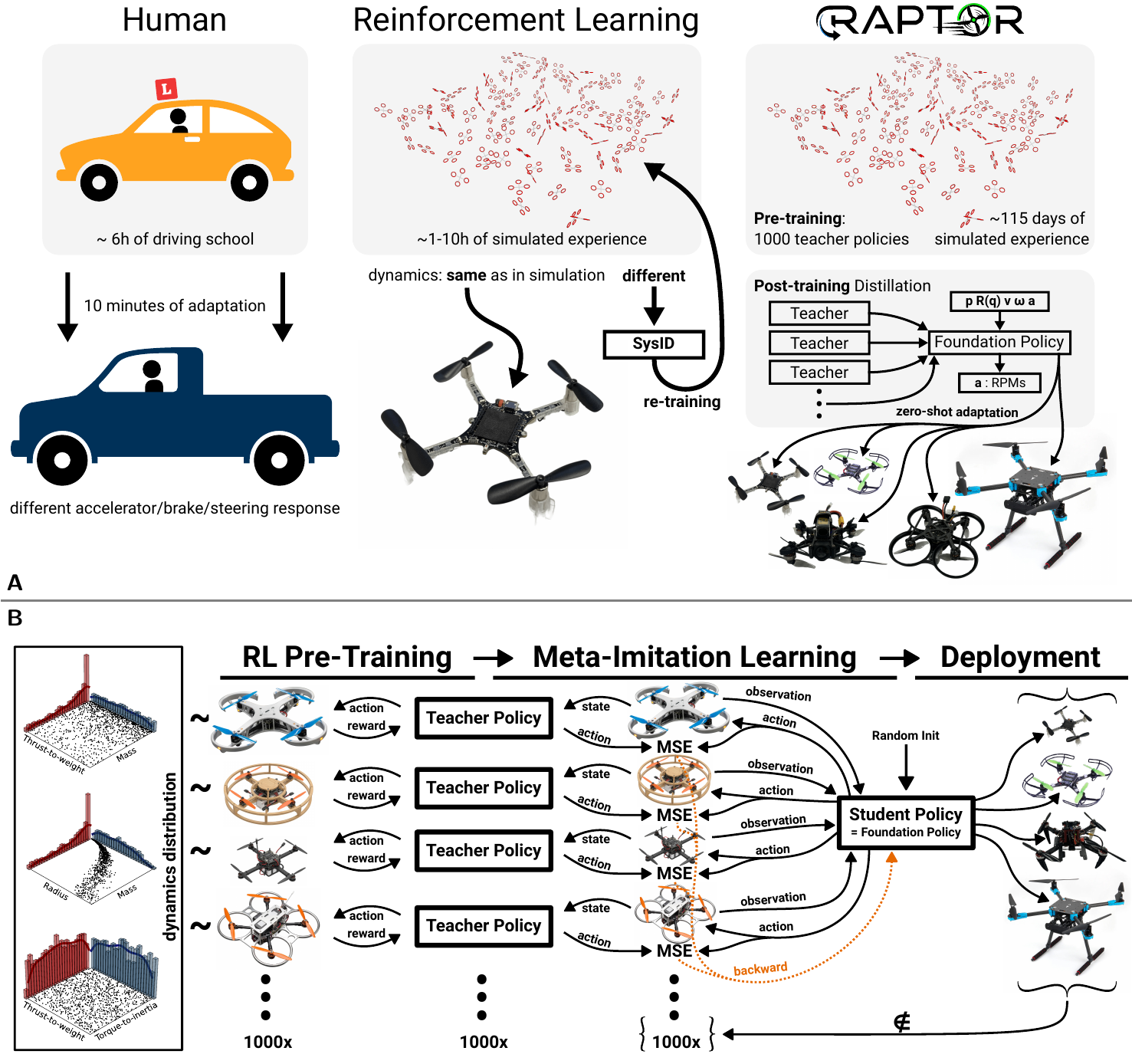} 
	\caption{\textbf{(A) Motivation.} Comparison of the adaptation capabilities of humans, contemporary RL-based control policies, and our RAPTOR method. \textbf{(B) The RAPTOR Method.} Overview of all stages of the RAPTOR architecture.}
	\label{fig:first-figure}
\end{figure}

% The first paragraph of any Science paper does NOT have a heading
% Nor is it indented
\noindent

\section*{Introduction}
\addcontentsline{toc}{section}{Introduction}
\label{sec:introduction}
Bearing Vertical Take-Off and Landing (VTOL) and hovering capabilities, Multirotor Aerial Vehicles (MAVs) have become a valuable platform for many real-world applications like package delivery \cite{guanrui2025Human-aware}, infrastructure inspection and maintenance \cite{aeroarms}, or search and rescue \cite{darpa-subt-cerberus}. In addition to the VTOL and hovering capabilities, MAVs can be built from cheap, mass-produced Commercial Off-The-Shelf (COTS) parts and can be scaled from light (tens of grams) and centimeter-sized to heavy (multiple kilograms) and meter-sized. This broad range allows for easy customization of the mechanical and electrical design for each particular application.

On the other hand, this variability also leads to challenges concerning the control of the platform. Changing the mechanical design and even simple modifications like switching the propeller or battery type often require the retuning of the cascade of classical controllers that is still most widely used today. Analogously, modern, nonlinear control approaches, like Model Predictive Control (MPC) or Reinforcement Learning, heavily rely on accurate system models.

Reinforcement Learning (RL) policies have recently gained widespread popularity for quadrotor control \cite{song2023reaching, kaufmann2023champion, ferede2025one, eschmann2023learning}. Although overcoming many challenges, the resulting policies either overfit a single dynamics model or rely on domain randomization without adaptation. Domain randomization of the dynamics parameters \cite{peng_sim--real_2018} is a powerful tool, but it forces fully Markovian (non-adaptive) policies to be conservative. Under domain randomization, the policy is incentivized to take conservative actions, for example, in a state/observation that is critical for some of the quadrotors (for example, due to thrust or angular acceleration limitations), the policy would have to take the action that saves these quadrotors, even if the current platform is agile enough to take the optimal action to continue the task.

In addition to overfitting a particular system model, in quadrotor racing, it has become standard to also overfit to a particular trajectory (such as a certain racetrack) \cite{loquercio2019deep, hanover2024autonomous}. This trend limits broad adoption of these methods for real-world use cases because even just changing the environment requires a full retraining of the policy.

Humans, unlike these methods, which need to be retrained from scratch for each particular platform and/or application, can adapt few-shot to new systems. An illustrative example is driving cars (see Figure \ref{fig:first-figure}A and Movie~S1). Initially, humans require extensive training to be able to control a car smoothly and robustly. But when they are driving a new, unseen car, they can adapt quickly. The steering, brake, and accelerator response might be quite different from the ones they have experienced before, but they usually only require a handful of tries to adapt.

With this work, we aim to devise a control policy that can adapt to unseen system dynamics using only minimal data, similar to humans. We are inspired by the recent progress on foundation models in the vision \cite{radford2021learning} and language domains \cite{brown2020language}. The two main premises of foundation models are:
\begin{enumerate}
\item \textbf{Broad Distribution}: Foundation models are trained on such a broad distribution of data that most expected queries at deployment time will be in-distribution with respect to the training data distribution. 
\item \textbf{In-Context Learning}: Foundation models take causal sequences as input, facilitating in-context learning.
\end{enumerate}

Therefore, the goal is that our policy can quickly adapt to novel systems by interacting with them and using the context/sequence of high-frequency interactions to reason about the dynamics. This can also be seen as emergent implicit system identification. The objective is to control the quadrotor, that is, based on the observations, output low-level motor commands that achieve an objective (such as controlling the position). Since the optimal outputs are dependent on the system dynamics, the (recurrent) policy has to learn to implicitly identify the unobserved/latent dynamics variables on the fly (figuratively and literally).

We refer to this as emergent \textbf{implicit} system identification because it only needs to infer the parts of the quadrotor dynamics parameters that are relevant to the input/output behavior of the system. These relevant parts can, for example, be ratios like thrust-to-weight ratio, torque-to-inertia ratio, but also aggregates of motor delays, thrust curves, and so on. We never train any neural network to explicitly reconstruct system parameters. The only training objective is performing well in terms of the reward function. This fulfills the premise 2) of foundation models.

The premise 1) of foundation models, training on such a broad distribution that all conceivable inference queries are in-distribution, is covered by designing a very broad distribution over dynamics parameters that covers virtually all realistic quadrotors. This distribution is then used to sample quadrotors to train the aforementioned adaptive policy with emergent system identification.

The six main research questions are:
\begin{enumerate}
\item \textbf{Feasibility}: Can a recurrent, end-to-end neural network policy express the described behavior? 
\item What \textbf{size} (number of parameters) does the recurrent neural network policy require to express this behavior? Can it run in hard real-time at high frequencies when deployed on \textbf{small microcontrollers}? 
\item What \textbf{context window} is feasible? Recurrent neural networks are notoriously hard to train for sequences longer than $100-200$ steps. Will the policy forget the system dynamics after a short time? 
\item Does the policy \textbf{generalize} to unseen quadrotors that are in-distribution and out-of-distribution? 
\item How much \textbf{time} is required from activating the policy until it has gathered enough information to stably control the quadrotor? Is this feasible mid-flight, or would the quadrotor crash before the policy has identified the system properly? 
\item Is there a trade-off between agility and adaptability?
\end{enumerate}

We tackle the question of feasibility 1) by devising a method to train such a foundation policy for quadrotor control, implementing it, and testing it on a range of real-world quadrotors.

We tackle the size and inference speed question 2) by studying the scaling laws \cite{kaplan2020scaling} in the student policy and by deploying the final foundation policy directly onto the microcontrollers of even the tiniest quadrotors.

We tackle the context window size question 3) by testing the context window extrapolation beyond the trained size.

We tackle the generalization question 4) by testing the policy on $9$ unseen but in-distribution (in terms of thrust-to-weight ratio, torque-to-inertia ratio, motor delays, and thrust curves) quadrotors in the real world. We also study the out-of-distribution performance by testing the foundation policy (a) on a quadrotor with a flexible frame (making it a total of 10 real quadrotors), (b) by installing four different propellers (2- and 3-blade), (c) by hitting it with a tool during flight, and (d) by testing it with a quadrotor that has thrust-to-weight ratio $> 2\times$ higher than the highest one experienced during training.

We tackle the question of the number of timesteps required to infer the system dynamics 5) by studying activation response trajectories, where the policy is activated in mid-air. Here, the policy needs to probe the system and infer the dynamics of it rapidly to restore or maintain stable flight.

We tackle question 6) about the agility-adaptability trade-off by testing the resulting foundation policy on the task of tracking trajectories from quasi-static to highly dynamic.

Answering these research questions, we provide the following five main contributions:
\begin{enumerate}
\item \textbf{RAPTOR} (\textbf{R}eal-time \textbf{A}daptive \textbf{P}olicy \textbf{T}hrough \textbf{O}nline \textbf{R}easoning): A method to train an end-to-end foundation policy for quadrotor control that can adapt to virtually any quadrotor platform zero-shot. This method consists of:
\begin{itemize}
\item Design of a distribution over dynamics parameters that resemble physically plausible quadrotors.
\item Our proposed distillation method called Meta-Imitation Learning that condenses the behavior of $1000$ Markovian teacher policies into a single adaptive/non-Markovian student policy.
\item A formal derivation of the design of the RAPTOR architecture. 
\end{itemize}
\item We contribute a highly efficient, open-source implementation of the aforementioned method that allows to reproduce our results even with resource-constrained, consumer-grade hardware. 
\item We study the scaling laws of the Meta-Imitation Learning process. 
\item We conduct extensive experiments (indoor and outdoor) across $10$ quadrotor platforms with different flight controller setups to validate that the foundation policy resulting from our proposed method answers the stated research questions and attains the stated goals. 
\item We provide robust and simple means to use our resulting foundation policy for quadrotor control in different flight controller firmwares as well as simulation environments. This facilitates the reproducibility of our results and simplifies its usage as a baseline in future works of the community.
\end{enumerate}

Although there have been many works on neural-network-based quadrotor control, most rely on lower-level controllers by, for example, outputting collective thrust and body-rate (CTBR) setpoints and, hence, are not fully end-to-end \cite{kaufmann_benchmark_2022, song2023reaching, kaufmann2023champion, zhang2024proxfly, heeg2024learning, xing2024multi}. But recently there have also been works investigating full neural-network-based end-to-end control \cite{gronauer2022using, ferede2024end, ferede2024end2, balandi2025acceleration, hegre2025neural}. Although these approaches have attained comparable performance to classical control schemes, the control policies are each highly optimized for a single quadrotor. Changing the quadrotor requires system identification to adjust the dynamics parameters of the simulator and a full retraining of the policy. This shortcoming is being tackled by the community right now, with works that investigate better adaptability and/or generalization of neural-network policies to multiple quadrotors. In particular, \cite{dingqizhang-extreme} and \cite{ferede2025one} are the most related works to our approach.

In \cite{ferede2025one}, the authors train a single neural-network policy that can be deployed onto two different quadrotors with thrust-to-weight ratios of $\approx 5.8$ and $\approx 11.0$, respectively. The authors demonstrate impressive agility for racing through gates. The main difference to our approach is that their policy is Markovian/stateless, whereas ours is adaptive.

Compared to \cite{ferede2025one}, \cite{dingqizhang-extreme} is more related to our approach because the method intends to train a policy that can adapt to different quadrotors. In \cite{dingqizhang-extreme}, the authors show deployment of their adaptive policy to two relatively similar quadrotors with thrust-to-weight ratios of $3.23$ and $3.62$, respectively. The biggest differences to our work are that their approach is not end-to-end. A high-level controller that outputs collective thrusts and body-rates (CTBR) is required. Their adaptive policy receives these CTBR setpoints as an input and is not concerned with rotational mechanics. Our foundation policy is a full position controller and hence, in contrast to \cite{dingqizhang-extreme}, also understands the tilt required to build up linear velocity and execute translations as well as the angular velocities required to execute a particular tilt. Our policy covers these major non-linear transfer functions whereas their policy only covers the angular rate (and thrust control) that is commonly implemented by a simple PD controller.
Additionally, our policy architecture is simpler because it does not require training two encoders that map into the same latent space. Furthermore, our policy is vastly lighter, requiring only $2084$ parameters whereas theirs requires $114872$ parameters ($55\times$ larger), even though our policy covers more complex behavior. Because their policy is so computationally intensive, we cannot run it on any of the $10$ quadrotors we are using for testing without modifying the hardware to include a more powerful inference computer.
Finally, the authors of \cite{dingqizhang-extreme} do not publish their full training code, making it hard to exactly replicate their results. In contrast, we publish the full training and inference code in an easy and future-proof way. Additionally, readers can interact with the foundation policy through the web app at https://raptor.rl.tools. 

Beyond quadrotor control, distilling the behavior of a privileged teacher into a partially observable student has become increasingly popular for robotic control tasks. In \cite{chen2020learning}, this idea is referred to as ``Learning by Cheating'', and a policy for driving cars is distilled from a teacher that is based on imitation learning with dense states from expert demonstrations into a student that only receives visual inputs. Compared to our approach, the inference is not performed by integrating information over time, but by extracting the remaining information after a lossy projection (ego-view). In \cite{doi:10.1126/scirobotics.abc5986} and \cite{KumarA-RSS-21}, the authors distill privileged teachers that can observe details about the terrain into student policies that only receive proprioceptive observations. Like in \cite{dingqizhang-extreme}, the latter uses an environment encoder and shares the control head between teacher and student. Like in our work, the students are sequence models. Although distilling a privileged teacher into a partially observable student appears like a simple concept at a high level, there are many nuances and design decisions, many of which have been studied recently in \cite{pmlr-v283-paluch25a}. The authors show the effectiveness of a GRU-based adaptive policy and thoroughly ablate many important aspects, but only demonstrate it using a cartpole, which is a relatively simple system compared to the quadrupeds and quadrotors used in this and the other referenced works. In contrast to all other works, we do not restrict our method to a single teacher. This circumvents well-known challenges in multi-task RL, where ``negative transfer'' is often observed \cite{pmlr-v100-yu20a, NEURIPS2020_3fe78a8a}, and additionally makes the pre-training stage embarrassingly parallel, which suits modern hardware well.
 
\section*{Results}
\addcontentsline{toc}{section}{Results}
Our results show that our method leads to a robust and highly adaptable foundation policy for quadrotor control. In the following, we first discuss insights from the training (pre- and post-training) using our method, as well as results from deploying the single foundation policy to multiple real-world platforms under different conditions.
\subsection*{Training}
\addcontentsline{toc}{subsection}{Training}
In the following, we describe the training results. The methods used to attain these results are described in the Materials and Methods section. As described in the Materials and Methods section and Figure \ref{fig:first-figure}B, our training method separates into a pre-training phase and a post-training phase.

\subsubsection*{Pre-Training}
\addcontentsline{toc}{subsection}{Pre-Training}
In the pre-training phase, we train $1000$ teacher policies that are each specialized for a single quadrotor. The quadrotors differ in their dynamics parameters, which are sampled from the distribution described in the Domain Randomization subsection.

We find that the pre-training method described in the Pre-Training subsection of the Materials and Methods section is remarkably robust and that all $1000$ RL training runs reliably converge to good policies, all of which can be used for the downstream Meta-Imitation Learning. This is surprising because RL training loops are known to be unstable and hence it is common to run the training with numerous seeds and then cherry-pick the best final policy \cite{henderson2018deep, wu2017scalable, van2016learning, lewis2019much, clary2019let, agarwal2021deep}.

In contrast, we can use the same initial seed in all $1000$ cases, and we do not need to cherry-pick a different seed for each quadrotor. Note that our training runs are deterministic, that is, a training run with the same seed always gives identical results (on the same computer). 
We attribute this stability to a combination of three main factors:
\FloatBarrier
\begin{enumerate}
\item \textbf{Overparameterization:} Since we do not need to deploy the teacher policies onto constrained hardware, we can use much larger ($\approx 55\times$) neural networks. This overparameterization is known to improve the loss landscape and stabilize training \cite{belkin2019reconciling}. 
\item \textbf{Reward Function:} We tuned the reward function for reliable training, particularly the survival bonus and the termination penalty. 
\item \textbf{Off-Policy RL:} We use an off-policy RL algorithm (SAC) because our experience matches the literature in that on-policy methods like PPO are usually more unstable and prone to local minima \cite{molchanov_sim--multi-real_2019}. We find off-policy methods to be particularly robust when the replay buffer retains all interaction steps and when combined with the previous two points.
\end{enumerate}
\FloatBarrier

\begin{figure}[t]
	\centering
	\includegraphics[width=1.0\textwidth]{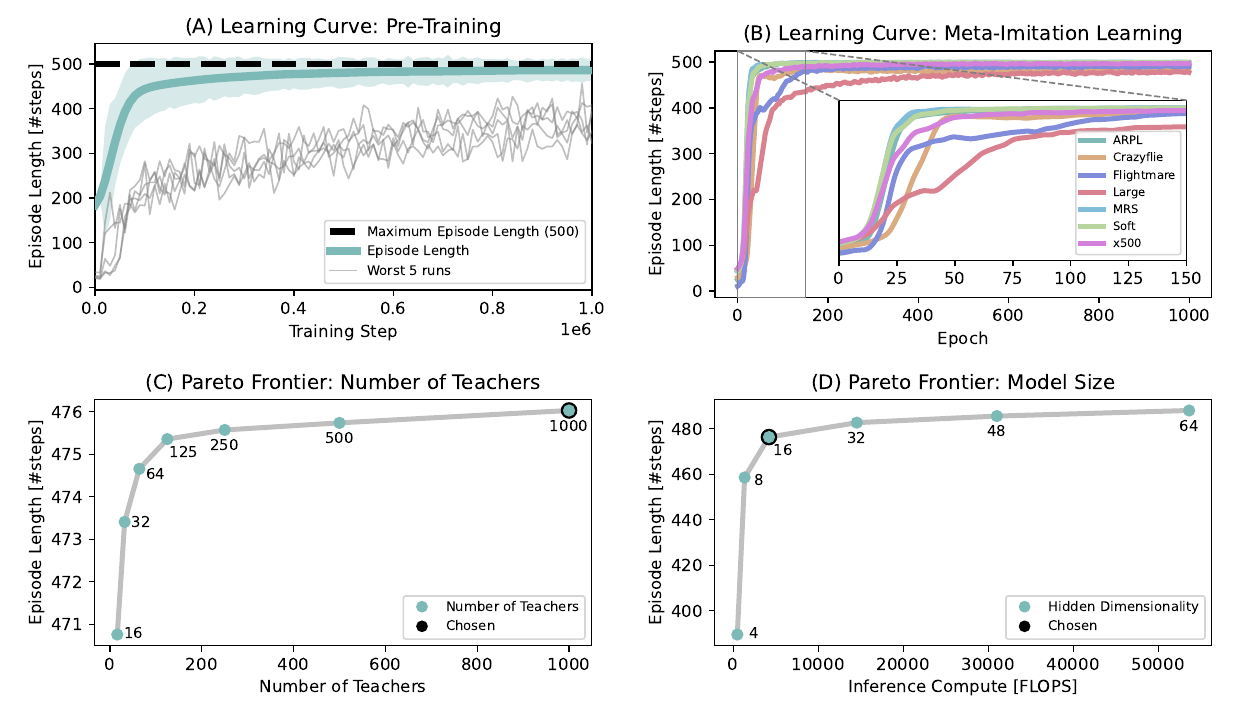}
	\caption{\textbf{Training Results.} \textbf{(A)} shows the pre-training learning curve, \textbf{(B)} shows the meta-imitation learning curve where the policy is evaluated using a validation set of $7$ quadrotors that are not seen during training, \textbf{(C)} shows the Pareto frontier between performance and number of teachers, and  \textbf{(D)} shows the Pareto frontier between performance and student/foundation policy size.}
	\label{fig:learning_curves_scaling_laws}
\end{figure}

If these conditions are met, it is just a matter of training long enough to see all training runs converge to a good policy (see Figure~\ref{fig:learning_curves_scaling_laws}A).
We also release the supplementary dataset Data S1 of $1000$ dynamics parameters and the $1000$ trained policies for each of them for further study.

The aggregated learning curve resulting from the pre-training phase can be seen in Figure \ref{fig:learning_curves_scaling_laws}A. We use the episode length as the metric here because it is more robust to the variation in quadrotor dynamics than the return. More agile quadrotors, for example, achieve much higher returns than less agile quadrotors, even if both are controlled by their respective optimal policy (with respect to the same reward function).

We find that the policies already achieve good behavior after only $100000$ steps of training, which is signified by the majority of policies reaching episode lengths of $400$ steps or more. We observe that the main flight capabilities are attained in the first $100000$ steps, and that subsequent steps refine the behavior when starting in challenging initial conditions (large tilt, large linear velocity, position close to the termination boundary, etc.) as shown in the episode lengths converging to the maximum limit. We also observe that, in the later stages of the pre-training, the steady-state performance (which is not well expressed in the episode length) is also still improving.

Since, in contrast to inference, the pre-training only incurs a one-time computational cost, we decide to train each teacher policy for $1$ M steps. Each training run takes $31$ min on a single core (consumer laptop, AMD Ryzen 9 7945HX, $16$ cores). Hence, the full pre-training takes about $34$ h on a single consumer laptop. Please refer to the Pre-Training subsection in the Materials and Methods section for details on why and how the pre-training can be parallelized and sped up.

\subsubsection*{Meta-Imitation Learning}
\addcontentsline{toc}{subsubsection}{Meta-Imitation Learning}
During the Meta-Imitation Learning phase, we observe that the student quickly learns to imitate any of the $1000$ teachers. In Figure \ref{fig:learning_curves_scaling_laws}B, we plot the test performance of the foundation policy when controlling $7$ unseen quadrotors in simulation. The set of $7$ quadrotors consists of $5$ quadrotors listed in Figure \ref{fig:quadrotor-table} that we deploy the foundation policy to and that we have accurate system parameters for (which is not the case for the remaining quadrotors in that table). Furthermore, we include two more quadrotors ("MRS" and "Large") that we have accurate system parameters for (from \cite{baca2021mrs} and \cite{eschmann2024datadrivenidentificationquadrotorssubject}, respectively).

None of these test-set quadrotors appear in the $1000$ pre-training quadrotors. We observe that the acquisition of good flying behavior follows a similar trajectory as a function of the epoch. For most quadrotors, the foundation policy achieves good performance early, but for the "Flightmare" and especially the "Large" model, we see continued improvement by training for longer. Eventually, after about $1000$ epochs, we observe convergence and good performance on all $7$ platforms, which span a large range of quadrotor dynamics, from agile to non-agile, lightweight to heavy, and more.

In Figure \ref{fig:learning_curves_scaling_laws}C, we study the performance of the resulting foundation policy when using the RAPTOR framework with different numbers of pre-training teachers/quadrotors. Here, the aggregate performance is shown across all $7$ test quadrotors, and we observe that, even for a low number of teachers, the foundation policy is able to stabilize/control its position from various and challenging initial states without crashing.

As mentioned before, the episode length is an imperfect metric for the full performance of the policy, but, nevertheless, it is more suitable than the return, which strongly varies in scale, even just between the $7$ test-set quadrotors. With a training time of $1.9$ h (same AMD Ryzen 9 7945HX laptop), Meta-Imitation Learning is vastly less computationally intensive than the pre-training ($1.9$ h vs. $1000\times31$ min).

The training part of the RAPTOR framework only incurs a one-time cost compared to the deployment part, which is more resource-constrained and hence more sensitive computationally. Hence, we decide to employ all $1000$ teachers in the main configuration of the RAPTOR framework.

Furthermore, and with the aforementioned resource constraints in mind, we also study the scaling behavior in the model size. In Figure \ref{fig:learning_curves_scaling_laws}D, we can see the Pareto frontier between inference compute in terms of Floating Point Operations (FLOPs) and performance in terms of episode length. Since, in this case, a larger model size actually entails continued computational costs during deployment, we chose a relatively small model with a hidden dimensionality of $16$.

This small size allows the deployment to even the tiniest microcontrollers. Even on the most resource-constrained quadrotors, the resulting foundation policy requires $< 10\%$ of the available computational power, leaving plenty of capacity for state-estimation, communication, and other duties of the flight controller.

Hence, going forward (and in Figure \ref{fig:learning_curves_scaling_laws}A/B), we set the number of teachers to $1000$ and the size of the hidden dimension in the policy to $16$.

\subsection*{Emergent System Identification}
\addcontentsline{toc}{subsection}{Emergent System Identification}
\label{sec:results-training-meta-imitation-learning}

\begin{figure}
	\centering
	\includegraphics[width=1.0\textwidth]{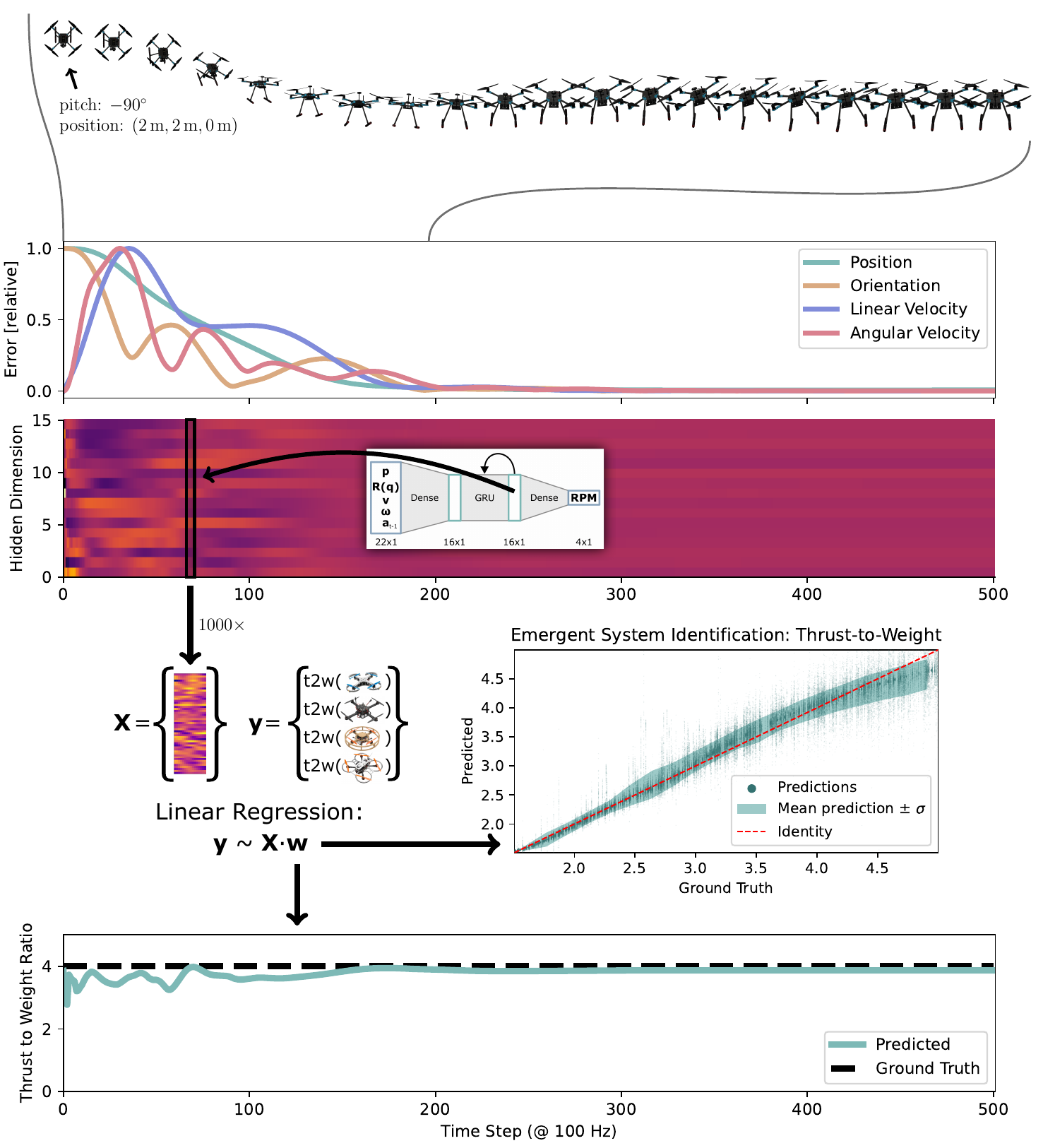}
	\caption{\textbf{Inference Results.} Here we show a recovery of a simulated quadrotor from an adverse initial condition using the trained foundation policy. We show the latent state of the policy throughout the trajectory and test if it performs emergent/implicit system identification by training a linear probe.}
	\label{fig:trajectory-plot-emergent-sysid}
\end{figure}

To answer the stated research questions, we also study the behavior of the foundation policy resulting from Meta-Imitation Learning. We find that it indeed exhibits in-context learning as can be seen in Figure \ref{fig:trajectory-plot-emergent-sysid}. Here we start a quadrotor in a state where it is displaced by $2$ m from the target position in $x$ and $y$ as well as pitched backwards by $90^\circ$ (no linear or angular velocity). First of all, we can see that the policy successfully manages to recover from this initial state and to reduce the distances towards the target state in position, orientation, linear and angular velocity.

Additionally, we plot the trajectory of the hidden state and investigate if it learns something about the dynamics of the quadrotor it is interacting with. We know the ground-truth dynamics parameters of the $1000$ quadrotors used for pre-training, hence we can train a linear probe \cite{alain2016understanding} to predict, for example, the thrust-to-weight ratio. Linear probing has become a standard test to evaluate if foundation models learn good representations \cite{radford2021learning,dosovitskiy2020image,oquab2023dinov2}.

As can be seen from the regression plot in Figure \ref{fig:trajectory-plot-emergent-sysid}, just a linear probe can predict the thrust-to-weight ratio very well. For this experiment, we collect $100$ new episodes (neither seen during Pre-Training nor Meta-Imitation Learning) from random initial states for each of the $1000$ sampled quadrotors. We use an $80\%$-$20\%$ train-test split and the linear model achieves a Mean-Squared Error (MSE) of $0.047$ and an $R^2$ of $0.949$, substantially reducing the a priori uncertainty about the thrust-to-weight ratio. This shows that Meta-Imitation Learning leads to emergent implicit system identification in the latent space of the foundation policy.

Although quickly going into the right range, we can also see that the estimate is improving over time, showing the in-context learning of the policy.

\subsection*{Deployment}
\addcontentsline{toc}{subsection}{Deployment}

\begin{figure}
	\centering
	\includegraphics[width=1.0\textwidth]{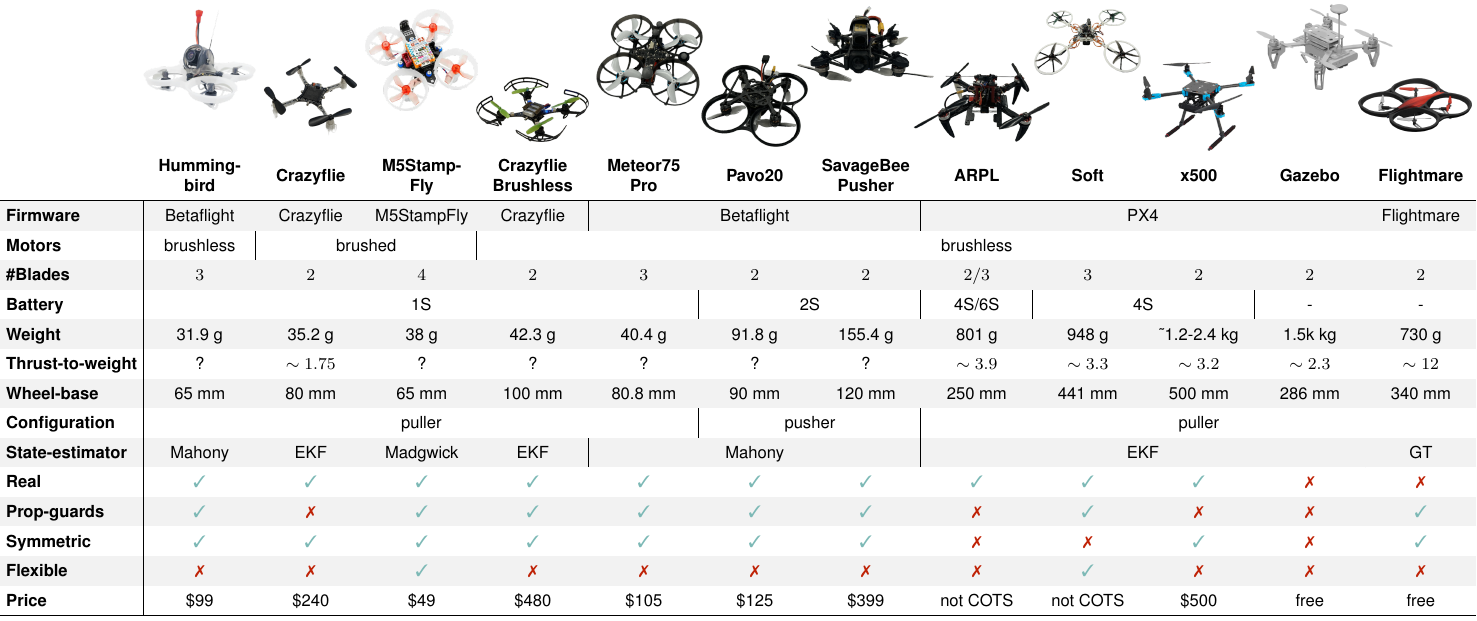}
	\caption{\textbf{Test Quadrotors.} A diverse set of $10$ real and $2$ simulated quadrotors that we use in the experiments.}
	\label{fig:quadrotor-table}
\end{figure}

To show the robustness and adaptability of the foundation policy that results from applying the RAPTOR framework, we deploy the foundation policy onto $10$ different real quadrotors and $2$ different simulators (see Figure \ref{fig:quadrotor-table}) showing Simulation-to-Reality (Sim2Real) and Simulation-to-Simulation (Sim2Sim) transfer. We aim to strain the adaptation capabilities of the foundation policy as much as possible by testing across a wide range of parameters:
\begin{enumerate}
\item A quantitatively wide range of parameters: weight ($31.9$ g - $2.4$ kg), size ($65$ mm - $500$ mm), and thrust-to-weight ($\approx 1.75$ - $12$). 
\item A qualitatively diverse set of features: flight controller (PX4, Betaflight, Crazyflie, M5StampFly), state estimator (EKF, Mahony, Madgwick), motor type (brushed and brushless), flexible frame, and mixing two- and three-blade propellers.
\end{enumerate}

Many of these quantities are (far) out-of-distribution, like a thrust-to-weight ratio of $12$ ($\leq 5$ in training), a flexible frame (only rigid during training), and observations from a state estimator (Ground Truth during training).

This shows that our proposed RAPTOR framework actually produces a policy that not only generalizes to quadrotors that are in the training distribution (see Domain Randomization subsection) but also out-of-distribution (OOD).

Contrary to popular belief, and supporting the results in \cite{eschmann2023learning}, we find that the simulation-to-reality transfer of end-to-end neural network policies is actually not hampered by the mismatch in the dynamics model, especially because the parameters can be relatively easily and accurately determined using \cite{eschmann2024datadrivenidentificationquadrotorssubject}. We find that qualitative differences matter much more. Especially for stateful policies (like the RNN used in the RAPTOR foundation policy), implementation details in the firmware that lead to delays and other artifacts have a strong influence on the simulation-to-reality deployment.

We find that the foundation policy works robustly on all platforms, but we also observe low-frequency z-axis oscillations in some of the non-EKF-based quadrotors. In the case of the Mahony \cite{mahony2008nonlinear} and Madgwick \cite{madgwick2010efficient} filters, only the orientation is estimated by the filter and the velocity is directly fed from the motion capture system.

We hypothesize that this leads to the z-axis oscillations due to communication delays.
We can reproduce the z-axis oscillations in simulation across quadrotors of different dynamics parameters by inducing a linear velocity delay of $10-30$ ms.

Intuitively, it makes sense that the foundation policy heavily relies on changes in the linear velocity to estimate the acceleration that is caused by the series of actions it produced previously. The policy can use these observations to estimate dynamics parameters like the thrust-to-weight ratio. Please refer to the Supplementary Materials \cite{methods} for the mitigation we implemented for this.

\subsubsection*{Trajectory Tracking}
\addcontentsline{toc}{subsubsection}{Trajectory Tracking}
\label{sec:results-deployment-trajectory-tracking}

\begin{figure}
	\centering
	\includegraphics[width=0.95\textwidth]{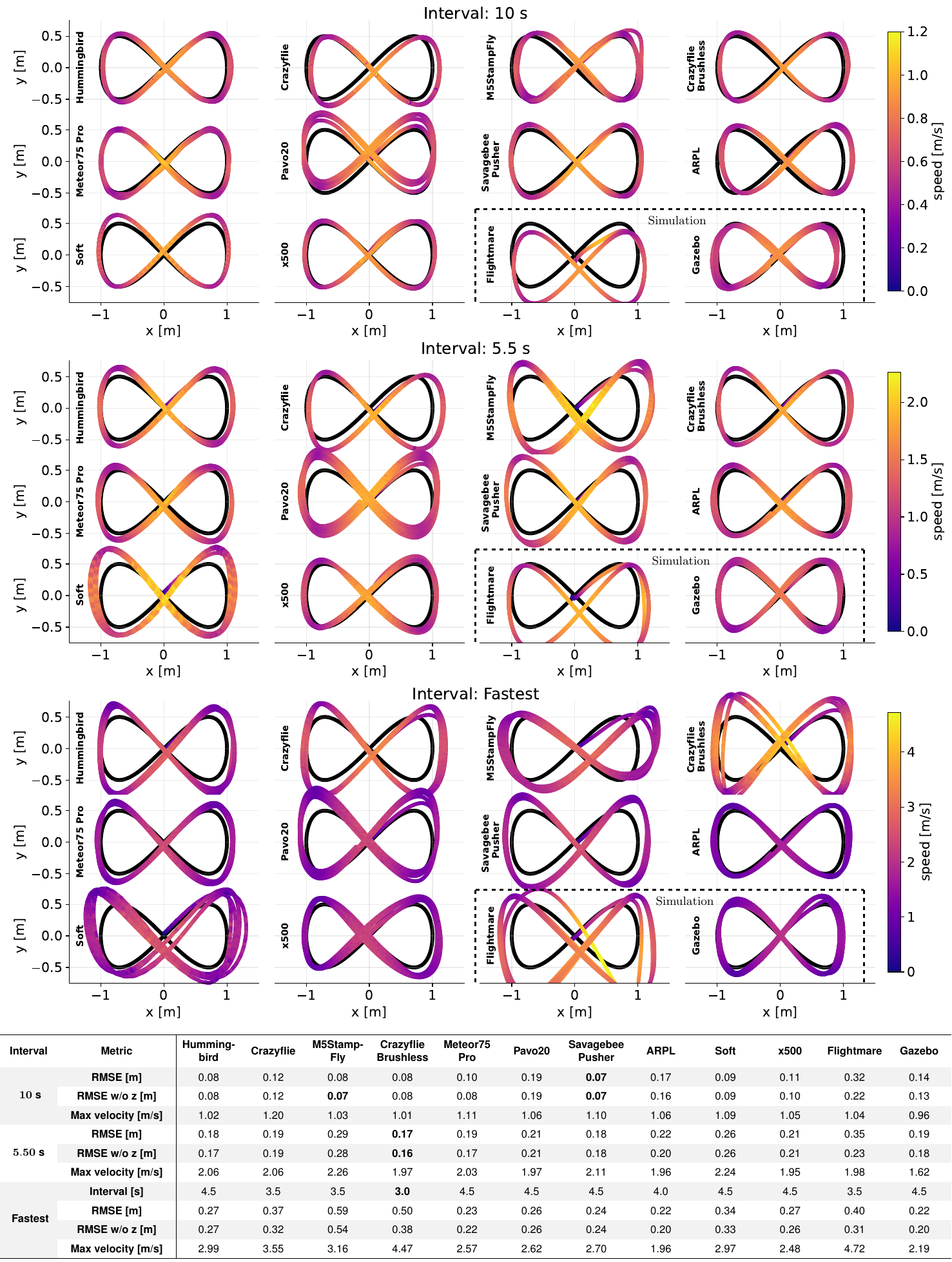}
	\caption{\textbf{Trajectory Tracking Results.} Trajectory tracking results of the $10$ real and $2$ simulation quadrotors.}
	\label{fig:trajectories}
\end{figure}

Besides position control (supporting "goto" workflows), trajectory tracking is an important task for real-world applications. Despite only training with random, relatively slow reference trajectories, we find that the resulting foundation policy is able to track figure-eight trajectories of varying agility well.

Figure \ref{fig:trajectories} and Movie~S2 show trajectory tracking of a Lissajous-based figure-eight trajectory at different intervals. We test tracking the $10$ s and $5.5$ s trajectories using the foundation policy on all $12$ quadrotors. Each of the shown trajectories constitutes $5$ consecutive full loops with an initial linear ramp-up time of $1$ s (from hovering).

Figure \ref{fig:trajectories} shows consistent real-world trajectory-tracking performance across repetitions. The Root Mean Square Error (RMSE) ranges between $0.07$~m and $0.19$~m (mean: $0.11$~m, standard deviation: $0.04$~m) for the $10$~s figure-eight trajectory and between $0.17$~m and $0.29$~m (mean: $0.20$~m, standard deviation: $0.04$~m) for the $5.5$~s figure-eight trajectory.

The tracking performance is in line with state-of-the-art policies (without trajectory lookahead, like the foundation policy) that are trained for a single quadrotor \cite{eschmann2023learning}. The dedicated policy that has been specifically trained for deployment on a Crazyflie in \cite{eschmann2023learning} reaches an RMSE tracking error of $0.17$ m and $0.15$ m (with and without the z-axis) for the identical $5.5$ s figure-eight trajectory. The foundation policy resulting from the RAPTOR framework reaches $0.19$ m and $0.19$ m, respectively, on the same platform. Although the tracking error of the foundation policy is slightly elevated compared to the dedicated policy, through online adaptation, the foundation policy can attain a similar performance on a plethora of other quadrotors as well. Note that our policy still performs better than the two other neural-network-based baselines evaluated in \cite{eschmann2023learning} which attain $0.23$ m / $0.21$ m \cite{gronauer2022using} and $0.25$ m / $0.24$ m \cite{molchanov_sim--multi-real_2019}, respectively.

From the RMSE difference between including the z-axis and excluding it, we can see that for all real quadrotors, the tracking error is mostly in the x-y plane. This shows that the foundation policy successfully adapts to the different thrust-to-weight ratios, battery levels, and other conditions, and adaptively cancels out the z-error.

The third cluster of trajectories in Figure \ref{fig:trajectories} shows the fastest trajectory that we tested for each quadrotor. This does not necessarily mean that this is the fastest trajectory supported by the foundation policy, because, to avoid crashes, we did not push all quadrotors (especially the larger ones) beyond their limits. For the Hummingbird, Crazyflie, M5StampFly, Crazyflie Brushless, and Meteor75 Pro, we did push them beyond their (or the foundation policy’s) limits and find that the shown trajectories are the most agile ones that can still be tracked with decent accuracy. When pushing, for example, the Hummingbird or the Meteor75 Pro further, they still remain stable, but they overshoot so much that the figure-eight becomes barely recognizable.

Based on this observation, we hypothesize that for agile trajectory tracking, the foundation policy is mainly bottlenecked by the lack of trajectory lookahead in the observations. This analysis is also based on prior lookahead-free works \cite{eschmann2023learning} and recent works ablating the inclusion of lookahead \cite{kunapuli2025levelin}.

As can be seen from the maximum velocity measurements, we push the foundation policy up to $3-4$ m/s in these indoor experiments. 

\subsubsection*{Outdoor Tests}
\addcontentsline{toc}{subsubsection}{Outdoor Tests}

Additionally, we conduct outdoor tests using the x500 platform. During these tests, there was a strong wind of about $7$ m/s, gusting up to $10$ m/s. We test trajectories resulting in linear velocities of up to $10$ m/s over ground and more than $15$ m/s relative to the wind. We did not see signs of instability at these speeds, and will investigate larger speeds in future work.

Furthermore, we equip the quadrotor with up to $1.2$ kg of payload (water-filled bottles), which, including the battery, is slightly above the specified maximum payload capability of the platform of $1.5$ kg. This setup leads to a take-off weight of $2.4$ kg, and the foundation policy is still able to control the quadrotor when hovering. With a payload of a single bottle ($600$ g), we can still track trajectories, as can be seen in the supplementary Movie~S3.

\subsubsection*{Disturbances}
\addcontentsline{toc}{subsubsection}{Disturbances}

\begin{figure}
	\centering
	\includegraphics[width=0.98\textwidth]{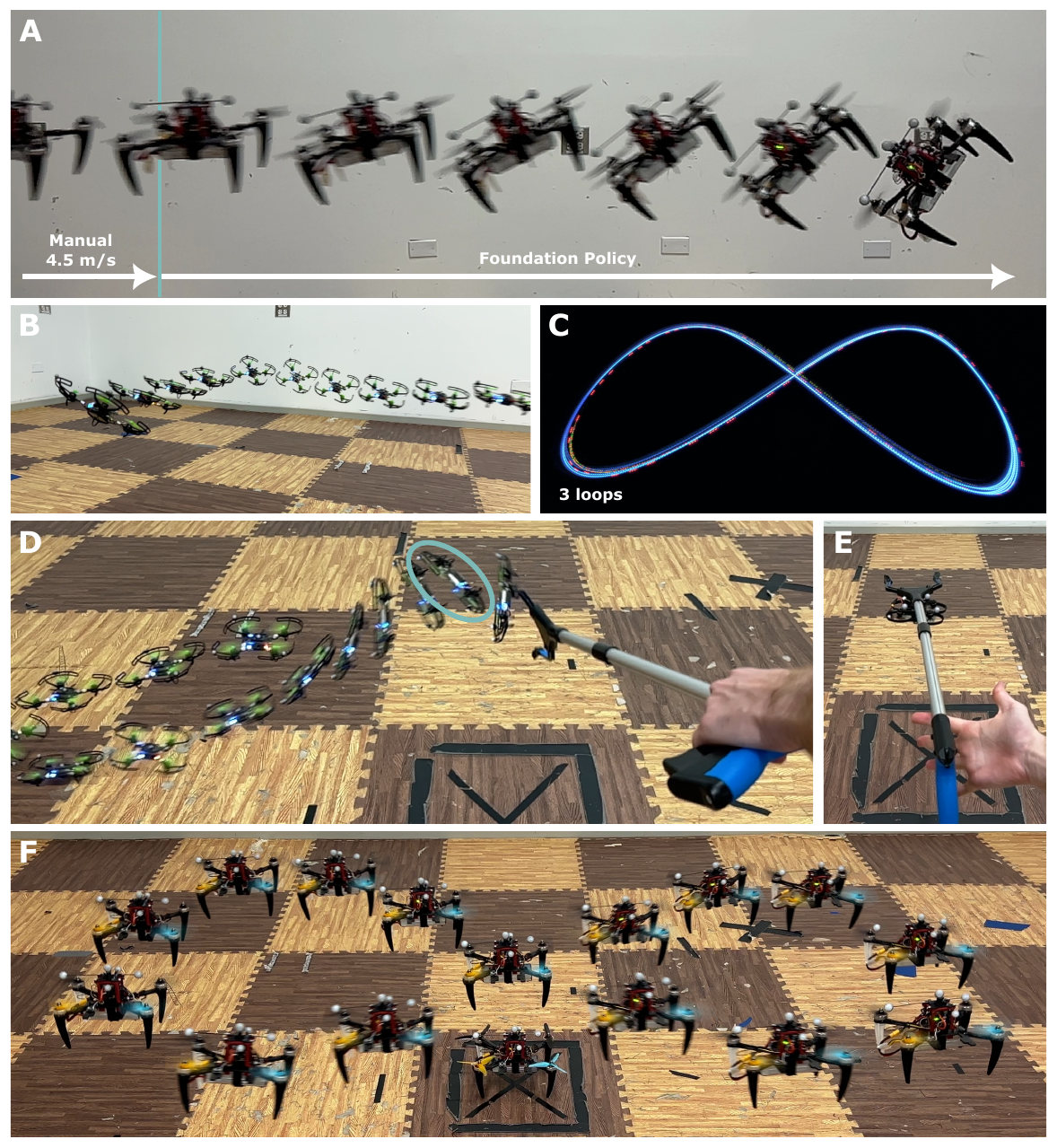}
	\caption{\textbf{RAPTOR Policy in Different Situations.} \textbf{(A)}: The foundation policy is activated in mid-air, starting with a linear velocity of $4.5$ m/s. \textbf{(B)}: Crazyflie Brushless tracking an agile trajectory. \textbf{(C)}: Long exposure photo of three consecutive loops of the Crazyflie Brushless trajectory. \textbf{(D)}: The foundation policy recovers from being poked and tilting $>90^\circ$. \textbf{(E)}: A tool is rested onto the quadrotor flown by the foundation policy. \textbf{(F)}: A quadrotor with four different propellers (2- and 3-blade) is tracking a trajectory using the foundation policy. }
	\label{fig:situations}
\end{figure}

We test a variety of disturbances, including hitting the quadrotors with a tool, resting the tool on top of the quadrotor while flying, as well as wind disturbances from a strong fan. Examples of these disturbances can be seen in Figure \ref{fig:situations} and Movies~S4 and S5. In Figure \ref{fig:situations}D and Movie~S4, we strongly hit the quadrotor (that starts out hovering) from the bottom, leading to a strong tilt in excess of $90^\circ$. While losing altitude, the foundation policy still manages to recover the quadrotor. In Figure \ref{fig:situations}E, we rest the tool on top of a flying quadrotor, and the foundation policy quickly adapts to the added weight and does not incur a steady-state altitude deficit. 

Additionally, we also test swapping out $1$, $2$, and $3$ propellers with random three-blade ones (the original ones are two-blade) and find that, in any of these configurations, the foundation policy still stably controls the vehicle and is even able to track trajectories (see Figure \ref{fig:situations}F and Movie~S6). Note that the distribution of quadrotors described in the Domain Randomization subsection covers only quadrotors with identical thrust curves for all four motors/propellers. Hence, neither the teacher policies nor the foundation policy has ever experienced differing thrust curves on the same quadrotor. Yet, we find that the foundation policy can generalize zero-shot outside of this distribution, and control the quadrotor well.

In the absence of disturbances, we find that the foundation policy yields a remarkably repeatable performance, as can be seen from the trajectories in Figure \ref{fig:trajectories}, which show $5$ consecutive full loops each. This also answers the posed research question about the trajectory length generalization of the foundation policy. During Meta-Imitation Learning, we train the policy with sequences of $500$ steps, corresponding to $5$ s of flight time. During inference, we test the foundation policy with $5$ consecutive iterations of, for example, a $10$ s, amounting to a trajectory length of $50$ s. This shows a $10\times$ context window size extrapolation.

In practice, we did not notice any limits on how long the policy can be activated at a time. We fly until the battery is empty (several minutes) many times over. Hence, we believe the resulting foundation policy can generalize to arbitrary trajectory lengths without degrading performance, despite only being trained on fixed-size trajectories.

Additionally, in Figure \ref{fig:situations}C, we can see the precise repeatability through a long-exposure photo of three full loops. The deviations between loops are barely recognizable. Figure \ref{fig:situations}B shows a chronophotograph of the same trajectory with the lights on.

Furthermore, we also test the recovery from aggressive initial states. An example of this is shown in Figure \ref{fig:situations}A and Movie~S7, where the quadrotor is accelerated to $4.5$ m/s using manual control and then the foundation policy is activated in mid-air. The policy is activated using a push button on the transmitter, and its goal (through offsetting of the observations) is to hover at the position where it was activated.

When it is activated, the hidden state is reset and the policy has to adapt zero-shot, in minimal time, to save the quadrotor.

Note that all of the previously described experiments (and more) are included in the supplementary video material, where these agile maneuvers and disturbances can be observed in motion. In Movie~S2, we also show flying $6$ different quadrotors (Hummingbird, Crazyflie, M5StampFly, Crazyflie Brushless, SavageBee Pusher, ARPL) from Figure \ref{fig:quadrotor-table} with 4 different firmwares and 4 different communication protocols at the same time in a tight indoor space. We find that, despite the turbulent flow created by the other quadrotors that are flying close by, the foundation policy still manages to stably control each of the quadrotors. We also accidentally fly a SavageBee Pusher ($155.4$ g) and a Crazyflie Brushless ($42.3$ g) directly beneath a hovering ARPL platform ($801$ g), and find that the foundation policy manages to quickly adapt and adjust to/recover from the disturbance. Furthermore, Movie~S8 contains yaw step-response tests.

\subsubsection*{Simulation}
\addcontentsline{toc}{subsubsection}{Simulation}
\label{sec:results-deployment-simulation}
To further test the out-of-distribution generalization beyond the aforementioned real-world tests including flexible frame, mixed propellers, and context window extrapolation, we test simulation-to-simulation transfer to the Flightmare simulator. This transfer is interesting, because the thrust-to-weight ratio of its default quadrotor  is $\approx 12$ (see Figure \ref{fig:quadrotor-table}) and hence $>2\times$ the upper limit of $5$ of our domain randomization range.
We find that the z-error is substantial, but the foundation policy still controls it robustly and can track agile trajectories. This shows the remarkable robustness and out-of-distribution generalization of the foundation policy resulting from the RAPTOR method.

\section*{Discussion}
\addcontentsline{toc}{section}{Discussion}

In our extensive experimental evaluations, we find that our proposed RAPTOR framework produces a highly robust and versatile foundation policy that can control a broad range of quadrotors in a large variety of situations. Compared to state-of-the-art solutions for end-to-end quadrotor control, our method is able to adapt to a wide range of physical platforms without re-training, through in-context learning. Even though existing methods rely on overfitting a control policy for each platform, our method is able to approximately match their performance while providing the flexibility of zero-shot adaptation and being substantially more light-weight, computationally. This allows, for example, deployment on platforms whose parameters have not yet been identified. Due to the minimal compute requirements of the tiny resulting policy, it can be deployed even on the smallest quadrotors.
Hence, we believe that our experiments validate the design choices in the RAPTOR architecture. 

Nevertheless, through experimentation in the real world, we find that there are many avenues for future research based on our proposed architecture, including the following $4$ directions:
\begin{enumerate}
\item Simulation-to-reality gap in flight controller firmware implementations: We find that a large part of the simulation-to-reality gap comes from delays in the flight controller firmware and infrastructure around the foundation policy as well as the state estimation. We see potential for introducing more domain randomization to simulate delays that appear during deployment, as well as moving to a fully end-to-end architecture by cutting out the state estimation and directly feeding IMU data into the policy. 
\item Over-reliance on linear velocity observations: We find that the foundation policy overindexes on the linear velocity observations to perform emergent system identification. In real-world deployments, there is usually a delay from the infrastructure (motion capture system) or other disturbing factors (such as drift in GPS). In future work, we believe this problem can be alleviated by adding more observation noise to the linear velocity observations and adding accelerometer measurements to the observation space. 
\item Limited domain randomization range: Although our distribution over quadrotor dynamics already covers a vast range of quadrotors (and arguably $>>50\%$ of quadrotors that are being used in the real world), our experiments showed that there are limits to the out-of-distribution generalization. We believe that the lower tracking performance on the Flightmare platform (thrust-to-weight ratio of $\approx 12$ vs. $\leq 5$ during training) can be alleviated by extending the domain randomization range because when limiting its thrust-to-weight ratio to $5$, we find it to yield excellent tracking performance. 
\item Trajectory tracking lookahead: We find that lookahead-free trajectory tracking is posing a limit and believe that integrating lookahead into the RAPTOR architecture can substantially improve the trajectory tracking performance.
\end{enumerate}

\section*{Materials and Methods}
\addcontentsline{toc}{section}{Materials and Methods}
\label{sec:materials-and-methods}
We formulate the quadrotor control problem as a Bayes Adaptive Partially Observable Markov Decision Process (BAPOMDP) \cite{ross2007bayes} defined by the tuple $(\mathcal{S}, \mathcal{S}_0, \mathcal{D}, \mathcal{A}, \mathcal{T}, \mathrm{r}, \mathcal{O}, \mathbf{o}, \gamma, \boldsymbol\Xi)$. $\mathcal{S}$ is the set of states $\mathbf{s} = \{\mathbf{p}, \mathbf{q}, \mathbf{v}, \boldsymbol\omega, \mathbf{a}_{t-1}, \boldsymbol\omega_m, \mathbf{f}_{\text{ext}}\}$ consisting of position, orientation, linear/angular velocity, previous action, motor states and a random external force, respectively. The previous action is included in the state because the reward function penalizes the change in action. $\mathcal{S}_0$ is the initial state distribution with position, orientation, linear/angular velocity uniformly sampled up to $10\cdot l_\text{arm}$, $90^\circ$, $1$ m/s, $1$ rad/s, respectively. With a probability of $10\%$, the initial state is overwritten with the target state (all zeros). The random force $\mathbf{f}_{\text{ext}}$ resembles, for example, wind disturbances and is sampled from a zero-mean normal distribution with a standard deviation derived from the sampled dynamics parameters (see supplementary materials \cite{methods}). $\mathcal{D}$ is the termination relation that includes all states outside $20\cdot l_\text{arm}$ m, $2$ m/s, $35$ rad/s (position, linear/angular velocity respectively). $\mathcal{A}$ is the set of actions $\mathbf{a} = \{\omega_{\text{sp}_0}, \omega_{\text{sp}_1}, \omega_{\text{sp}_2}, \omega_{\text{sp}_3}\}$ (individual motor commands). The transition probabilities $\mathcal{T}$ are defined as $\mathrm{p}(\mathbf{s}_{t+1} | \mathbf{s}_{t}, \mathbf{a}_t, \boldsymbol\Xi)$ and implemented by the L2F simulator \cite{eschmann2023learning}. The reward function  is deterministic:
\begin{align}
 \mathrm{r}(\mathbf{s}_{t}, \mathbf{a}_t, \mathbf{s}_{t+1}) = -\|\mathbf{p}\|_2 - 0.2 \cdot \arccos(1 - |q_z|) - \|\mathbf{a}_t - \mathbf{a}_{t-1}\|_2 + 1.5 - 100 \cdot \mathbf{1}[\mathrm{terminal}(\mathbf{s}_{t+1})]
\end{align}
We include the next state in the reward function to be able to inflict the termination penalty. The observation space $\mathcal{O}$ contains a subset of the state space $\mathbf{o} = \{\mathbf{p}, \mathbf{R}(\mathbf{q}),\mathbf{v}, \boldsymbol\omega, \mathbf{a}_{t-1}\}$, occluding the motor states $\boldsymbol{\omega_m}$ (not observable on most real-world platforms) and the external disturbance. $\gamma = 0.99$ is the discount factor and the domain parameters are collected in $\boldsymbol\Xi = \{m, l_{\text{arm}}, c_{f_0}, c_{f_1}, c_{f_2}, c_m, \mathbf{J} = \text{diag}(J_{xx}, J_{yy}, J_{zz}), T_{m\uparrow}, T_{m\downarrow}\}$ containing the mass, arm length, thrust-curve coefficients (zeroth, first and second order), moment coefficient, inertia matrix and rising/falling edge motor delays respectively. The difference between a BAPOMDP and a POMDP is that we can factor out the system parameters $\boldsymbol\Xi$ (which would have to be encoded into the state in a normal POMDP). This factorization allows us to implement the inductive bias that the parameters are only sampled once at the beginning of the episode and remain constant throughout.

\begin{figure}
        \centering
        \includegraphics[width=1.0\textwidth]{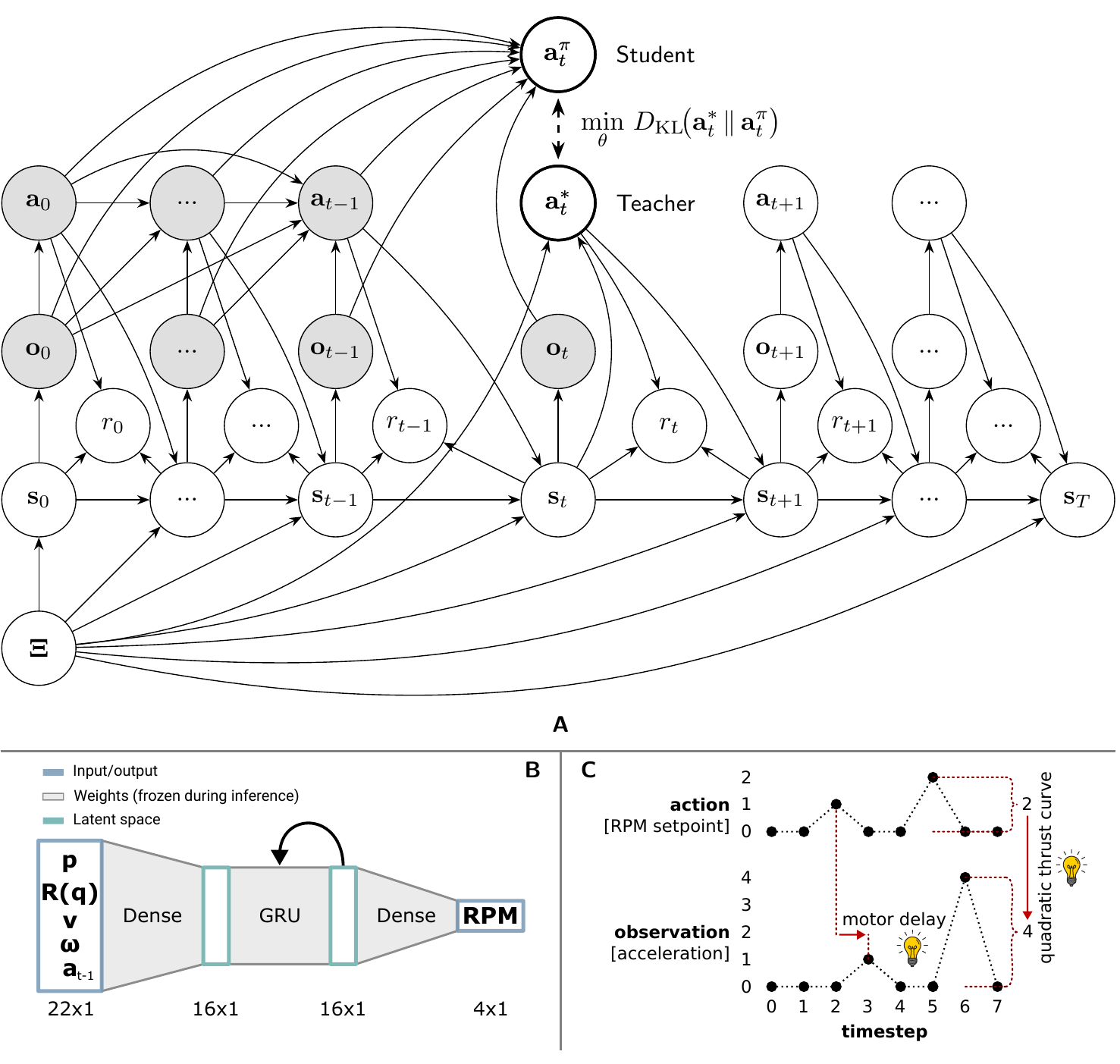}
        \caption{\textbf{(A)}: A probabilistic graphical model (Bayesian Network) of the dynamics and control of a random quadrotor. This formal model allows us to derive the RAPTOR architecture from probabilistic principles. \textbf{(B)}: Foundation policy network architecture. \textbf{(C)}: Illustration of inferring dynamics parameters by reasoning about the observed input/output behavior of the system.}
        \label{fig:probabilistic-model-arch-concept}
\end{figure}

Using this BAPOMDP framework, in the following, we provide a formal derivation of our method using a probabilistic graphical model \cite{koller2009probabilistic}.
The full model is shown in Figure \ref{fig:probabilistic-model-arch-concept}A. The goal is to model the decision-making at time $t$. At timestep $t$, $t$ previous observations and actions, as well as the current observation have been observed (shaded nodes). We explicitly place minimal assumptions on the policy that decides the previous actions, which shows by the previous actions being causally fully connected to all previous actions and observations. By causally fully connected, we mean that an action at timestep $t$ might depend on observations $\mathbf{o}_0, \cdots, \mathbf{o}_t$ and actions $\mathbf{a}_0, \cdots, \mathbf{a}_{t-1}$. Hence, the only assumption is that the policy generating the previous trajectory is causal. 

Since we want to maximize the expected discounted return (sum of rewards), we model the probability of some action $\mathbf{a}$ being the optimal action $\mathbf{a}_t^* := \mathbf{a}$. The optimal action is independent of previous states (and all other past random variables) given the current state $\mathbf{s}_t$ and the dynamics parameters $\boldsymbol\Xi$: 
\begin{equation}
\mathbf{a}_t \indep \mathbf{s}_0, \ldots, \mathbf{s}_{t-1} \mid \mathbf{s}_t, \boldsymbol\Xi
\end{equation}
Due to the forward-looking nature of maximizing the discounted future returns, for completeness, we assume that future actions are also optimal. In our proposed Meta-Imitation Learning method, this distribution over the optimal action $\mathbf{a}_t^*$ is approximated by a teacher policy that is trained using RL. In our method, we train $1000$ teacher policies for each of the $1000$ randomly sampled quadrotors/dynamics parameters $\boldsymbol\Xi$. This fits into the formal model in Figure \ref{fig:probabilistic-model-arch-concept}A because the practical, finite number of dynamics parameters/teacher policies can be viewed as a Monte Carlo approximation for the mixture model, where a unified teacher is conditioned on $\boldsymbol\Xi$. 

Since we want to train a foundation policy that can adapt to any realistic quadrotor and does not require knowledge of identified system parameters, and due to the motor states being unobservable on most quadrotor platforms, we can observe neither of the two direct ancestors $\mathbf{s}_t$ and $\boldsymbol\Xi$ of the optimal action. 

Using d-Separation \cite{pearl1988probabilistic}, an algorithm to prove conditional independencies in probabilistic graphical models, we can show that not observing $\mathbf{s}_t$ and $\boldsymbol\Xi$ makes the distribution over the optimal action $\mathbf{a}_t^*$ dependent on other variables that could carry information about them. The most direct example is $\mathbf{o}_t$. Since $\mathbf{o}_t$ is derived from $\mathbf{s}_t$, it can carry information about $\mathbf{s}_t$ and in our case it carries almost all the information about the state, apart from the motor speeds. Hence, when $\mathbf{o}_t$ is observed, this changes the distribution over the optimal action $\mathbf{a}_t^*$. The same argument holds for $\mathbf{o}_{t-1}$ and earlier observations because, for example, $\mathbf{o}_{t-1}$ carries information about $\mathbf{s}_{t-1}$ and $\mathbf{s}_{t-1}$ (in combination with the observable $\mathbf{a}_{t-1}$) carries all the information to infer the distribution over $\mathbf{s}_t$. This is because $\mathbf{s}_{t-1}$, $\mathbf{a}_{t-1}$, $r_{t-1}$, and $\boldsymbol\Xi$ form the causal Markov blanket \cite{pearl1988probabilistic} for $\mathbf{s}_t$. For the causal Markov blanket, random variables that lie in the future are removed. 
Therefore, using the d-Separation rules, we can see that the distribution over the optimal action $\mathbf{a}_t^*$ is dependent on all previous observations and actions. There is no set of observable variables that could form a Markov blanket and shield this dependency. This motivates our design decision that the foundation policy, which is the student policy from the Meta-Imitation Learning perspective, is dependent on all previous observations and actions. Intuitively, the dependence on all previous observations and actions makes sense because any of the individual observations, or, more likely, the combination of observations over time, contains information about the dynamics parameters $\boldsymbol\Xi$ and about the ground-truth state $\mathbf{s}$.

This is illustrated in Figure \ref{fig:trajectory-plot-emergent-sysid} and Figure \ref{fig:probabilistic-model-arch-concept}C where we show how the relation of observations over time carries information about the dynamics parameters. By looking at the observation-action history, the policy can observe that, for example, the effect of the motor commands is always delayed by one timestep. Note that this is a simplification by using a pure delay/dead time, whereas in reality the temporal relationship between motor commands and motor speeds is more closely modeled as a first-order low-pass filter. Furthermore, it can be observed that a motor command of $1$ corresponds to an observed acceleration of $1$, but a motor command of $2$ corresponds to an observed acceleration of $4$. Hence it can infer the curvature of the thrust curve. 

Therefore, from both the theory and intuition, we can conclude that the foundation policy should be able to take past observations and actions as the input and then output the predicted distribution over optimal actions as closely as possible. Therefore, we model the action distribution of the foundation policy (student in the Meta-Imitation Learning framework) with dependencies on all previous observations and actions in Figure \ref{fig:probabilistic-model-arch-concept}A. Even given the whole history of observations and actions, there might still be mutual information left between $\mathbf{s}_t$ and $\boldsymbol\Xi$. Hence, we cannot expect the distribution $\mathbf{a}_t^\pi$ to model the optimal action distribution $\mathbf{a}_t^*$ exactly. 

Instead, we want to model it as closely as possible and hence phrase the problem as variational inference, where we try to minimize the Kullback-Leibler (KL) divergence between the predicted optimal action distribution and the actual optimal action distribution. Note that due to the non-linearities in the system dynamics, the actual optimal action distribution is not tractable and that we approximate it by training expert teacher policies until convergence. 

Please refer to the Supplementary Materials \cite{methods} for a full derivation of the Mean-Squared Error (MSE) training objective from Maximum Likelihood Estimation (MLE).

Whereas the Bayesian Network in Figure \ref{fig:probabilistic-model-arch-concept}A is a formal/mathematical probabilistic model, Figure \ref{fig:first-figure}B shows our proposed practical method for modeling the various conditional probability distributions that constitute it:
\begin{enumerate}
\item \textbf{Dynamics Distribution/Domain Randomization}: This distribution implements the $\boldsymbol\Xi$ node. 
\item \textbf{RL Pre-Training}: We use RL to train teacher policies that act as an oracle for the optimal action distribution node $\mathbf{a}^*_t$. Since we train $1000$ specialized actors, the $\mathbf{a}^*_t$ node is implemented by a mixture policy, where the selected teacher is dependent on the dynamics parameters $\boldsymbol\Xi$, which is also well characterized by the Bayesian Network through the dependence of $\mathbf{a}^*_t$ on $\boldsymbol\Xi$. 
\item \textbf{Meta-Imitation Learning / Student Policy}: This models the partially observable action distribution node $\mathbf{a}^\pi_t$ and is implemented by a recurrent policy, which takes a history of observations and actions as input. 
\item \textbf{Deployment}: We deploy the foundation policy onto different, unseen, real-world quadrotors. We assume that the dynamics of most real quadrotors are in-distribution with respect to the distribution over dynamics parameters $\mathrm{p}(\boldsymbol\Xi)$.
\end{enumerate}
In the following, we describe these modules of the RAPTOR architecture in more detail.
\subsection*{Domain Randomization}
\addcontentsline{toc}{subsection}{Domain Randomization}
\label{sec:materials-and-methods-sampling-quadrotors}
We want the resulting foundation policy to be able to control a wide range of quadrotors. The RAPTOR philosophy is to employ radically wide domain randomization over $\boldsymbol\Xi$ and take advantage of emergent meta-learning \cite{openai_solving_2019} to produce a foundation policy that can adapt to unseen quadrotors zero-shot. To facilitate this, we need to design a distribution over realistic quadrotors that assigns a sufficient amount of probability mass to real-world quadrotors.

We are mainly concerned with the mass, geometric dimensions, inertia, thrust curves, torque coefficients, and motor delays, since these capture the most important parts of the quadrotor dynamics. These quantities are correlated in non-linear ways, making it intractable to directly formulate the joint distribution in analytical form.

Hence, we factorize the distribution based on physical properties.
By formulating the distribution in this factorized way, we can use efficient ancestral sampling to sample new quadrotors. Please refer to the supplementary Figure \ref{fig:sampling-quadrotors-graphical-model} for a graphical model corresponding to this factorization and ancestral sampling scheme. Due to this scheme, we prevent having to resort to heavy sampling mechanisms like Markov Chain Monte Carlo (MCMC), and can sample the root (shaded) quantities from simple, independent uniform distributions. Even though the marginal distributions are independent, the computed nodes are dependent on multiple inputs and are correlated in a physically plausible way.

Please refer to the Supplementary Materials \cite{methods} for the detailed equations that establish the physically plausible correlations.

\subsection*{Training Methodology}
\addcontentsline{toc}{subsection}{Training Methodology}
After establishing a realistic distribution over quadrotors, the question arises on how to devise a foundation control policy that can adapt to any one of them. We initially experimented using end-to-end RL with a single recurrent policy and critic, in the spirit of meta-RL \cite{wang2016learning,duan2016rl}, but we did not see signs of convergence, and the training was very time-intensive due to the sequential nature of training Recurrent Neural Networks (RNNs).

Since we knew that by combining \cite{eschmann2023learning} and \cite{eschmann2024datadrivenidentificationquadrotorssubject}, we can train good individual RL policies for a wide range of quadrotors, and due to the dependencies derived from the probabilistic model, we made the design choice to factorize the architecture into a pre-training and post-training/Meta-Imitation Learning stage. This architectural division is also inspired by the common practice of splitting the training of language and vision foundation models into pre- and post-training \cite{achiam2023gpt}.

\subsubsection*{Pre-Training}
\addcontentsline{toc}{subsubsection}{Pre-Training}
\label{sec:materials-and-methods-training-pre-training}

In the pre-training phase, we take $1000$ quadrotors sampled from the distribution described in the Domain Randomization subsection and train a dedicated expert policy for each of them by creating an independent MDP for each set of sampled domain parameters $\boldsymbol\Xi$. We also make the ground-truth states directly observable because this expert policy does not need to be deployed on hardware. Since we are not constrained by the deployment onto hardware, we can overparameterize it to aid the training \cite{belkin2019reconciling}. We use fully-connected three-layer neural networks with a hidden dimensionality of $64$, making each teacher policy $>3 \times$ larger than the condensed foundation policy in terms of parameters. The training pipeline is adapted from \cite{eschmann2023learning} with five modifications:
\begin{enumerate}
\item Switching from TD3 to SAC because we observed slightly more robust training dynamics in SAC. Note that we use an off-policy method because in our experience they are much more reliable than on-policy methods like PPO and, for example, do not require cherry-picking random seeds. The latter is of utmost importance because we do not have the capacity to supervise each of the $1000$ training runs, and hence rely on each of them finding a good policy without any adjustments to hyperparameters. 
\item Training for longer to ensure convergence for all quadrotors. 
\item Adjusting the reward function, adding a penalty for termination and for the action derivative. 
\item Removing the curriculum because we found that the changes to the reward function stabilize the training without the need for a curriculum. 
\item Ground-truth motor RPM states. The teacher policies are never deployed in reality, so instead of feeding a proprioceptive action history to account for the unobservable motor states as in \cite{eschmann2023learning}, the teachers can directly observe the ground-truth motor states. This also makes the actor-critic architecture symmetric.
\end{enumerate}

We do these modifications to trade off wall-clock training time for highly reliable training dynamics, and we found them to yield high-quality teacher policies in all $1000$ cases without cherry-picking random seeds and without requiring case-by-case modifications, even though the dynamics of the quadrotors vary drastically as described in the Domain Randomization subsection.

The teachers observe the states fully $\mathbf{o}_\text{teacher} = \{\mathbf{p}, \mathbf{R}(\mathbf{q}), \mathbf{v}, \boldsymbol\omega, \mathbf{a}_{t-1}, \boldsymbol\omega_m, \mathbf{f}_{\text{ext}}\}$.

We train using a position control objective where the quadrotor is initialized in a random state (for example, up to $90^\circ$ tilt) and the policy has to navigate it back to the origin with zero yaw while also minimizing linear/angular velocity and action changes.
To prepare the policy for trajectory tracking, we also train with the objective of tracking a relatively slow trajectory sampled from a second-order Langevin process. The Supplementary Materials \cite{methods} contain additional details about the motivation for and the implementation of the distribution over reference trajectories.

\subsubsection*{Meta-Imitation Learning}
\addcontentsline{toc}{subsubsection}{Meta-Imitation Learning}
\label{sec:materials-and-methods-meta-imitation-learning}

After training $1000$ teacher policies, we would like to distill all of their behaviors into a single student foundation policy. From the perspective of each teacher policy, there are no hidden/latent parameters because each teacher policy can assume that it always interacts with the same quadrotor $\boldsymbol\Xi$. However, in aggregate, and from the perspective of the student policy, the parameters of the quadrotor $\boldsymbol\Xi$ it is interacting with are not observable and need to be inferred as described in the Materials and Methods section. Due to the variable number of past steps (see Figure \ref{fig:probabilistic-model-arch-concept}A), we design a Gated Recurrent Unit (GRU) \cite{cho2014learning}-based foundation policy architecture as displayed in Figure \ref{fig:probabilistic-model-arch-concept}B.

The relatively small hidden dimensionality of $16$ is justified by the scaling experiments in the Meta-Imitation Learning subsection of the Results section. Due to the recurrence, the policy can theoretically "access" all the previous observations and actions.

We refer to our proposed algorithm as Meta-Imitation Learning because the student not only needs to learn to recreate the teachers' outputs from a different set of inputs/sensors, but also must learn to perform inference about the current MDP that it is acting in. Each quadrotor constitutes a separate MDP because the transition function varies based on the random dynamics parameters $\boldsymbol\Xi$. We currently only tackle the case where the dynamics parameters change, but in the future, we will also incorporate, for example, changes in the reward function.

\begin{figure}
        \centering
        \includegraphics[width=0.5\textwidth]{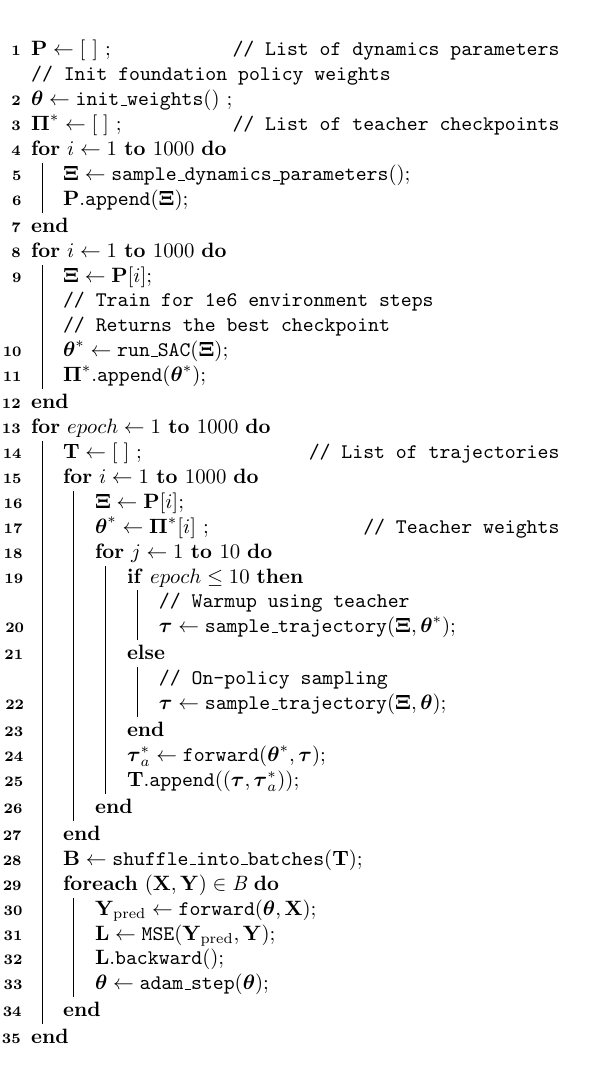}
        \caption{\textbf{Meta-Imitation Learning Algorithm.}}
        \label{fig:algorithm}
\end{figure}

Our proposed method is conceptually similar to the DAgger algorithm \cite{ross2011reduction} but differs in two key ways, firstly in the aforementioned requirement for meta-learning and secondly in that we perform on-policy data collection (after warm-up) and learning whereas DAgger is performing on-policy data collection but also uses off-policy data (the aggregated dataset) for learning. Figure \ref{fig:algorithm} shows our full proposed algorithm consisting of the sampling of $1000$ random quadrotors and the two main learning phases: pre- and post-training.

In the post-training phase, we distill the combined behaviors of the $1000$ teachers into the student foundation policy. We train it for $1000$ epochs and solely use on-policy data after a warm-up (using teacher rollouts) of $10$ epochs. The task of the student foundation policy, characterized by its parameters $\boldsymbol\theta$, is to predict the teachers' motor commands as closely as possible just based on the history of observations and (its own actions), without knowing the teacher or dynamics parameters at hand.

As shown in Figure \ref{fig:probabilistic-model-arch-concept}A, this forces the policy to infer the parameters of $\boldsymbol\Xi$ that are relevant for the input/output behavior of the system. This meta-learning using in-context reasoning is the central part of our proposed method. Additionally, we propose the use of on-policy imitation learning, where we neither use the actions of the teachers during rollout (in contrast to the $\beta$ trade-off in \cite{ross2011reduction}) nor use trajectories from past policies. We use on-policy imitation learning because full dataset aggregation as in DAgger \cite{ross2011reduction} is infeasible due to memory constraints, and we also find it to learn faster and better policies.

\subsection*{Computational Aspects}
\addcontentsline{toc}{subsection}{Computational Aspects}
\label{sec:materials-and-methods-computational-aspects}

Computationally, the separation into pre-training and post-training/meta-imitation learning is a major advantage of the RAPTOR framework. This decouples the time/compute-intensive pre-training and renders it "embarrassingly parallel" \cite{moler1986matrix}. This can be seen from Figure \ref{fig:first-figure}B, where the teacher training processes are independent. Hence, we can horizontally scale out the number of training processes over multiple processors and/or machines, linearly speeding the pre-training up to the ceiling, which is the $31$~min duration of each training run (at $1000$ cores in parallel). In contrast to LLM pre-training, where the gradients have to be communicated between all nodes at every training step, in RAPTOR, pre-training is entirely independent and communication/synchronization is only required after pre-training, in the distillation/meta-imitation learning phase, which is about three orders of magnitude less computationally intensive than the pre-training.

%%%%%%%%%%%%%%%% MAIN TEXT FIGURES %%%%%%%%%%%%%%%
\newpage

%%%%%%%%%%%%%%%% MAIN TEXT TABLES %%%%%%%%%%%%%%%

%%%%%%%%%%%%%%%% REFERENCES %%%%%%%%%%%%%%%

\clearpage % Clear all remaining figures and tables then start a new page

% The list of references goes after the main text and before the acknowledgements
% When preparing an initial submission, we recommend you use BibTeX, like this:
%
\bibliography{science_template} % for a file named science_template.bib
\bibliographystyle{sciencemag}

% After the paper has completed peer review and been revised ready for acceptance,
% you should comment out the lines above and copy-paste the contents of your .bbl
% file here instead. This will help ensure that our conversion software works correctly.
% Remember to re-run BibTeX first - check the timestamp!
%
% Example of the first three entries copy-pasted from science_template.bbl:
%
%\begin{thebibliography}{1}
%
%\bibitem{example}
%A.~N. {Author}, An example reference. \emph{Journal of Improbable Research}
%  \textbf{1}, 67 (2020).
%
%\bibitem{example2}
%F.~M. {Surname}, S.~{Author}, A second example. \emph{Interesting Research
%  Letters} \textbf{32}, 897 (2019).
%
%\bibitem{example_preprint}
%P.~{One}, P.~{Two}, P.~{Three}, {An unpublished preprint}. \emph{preprint}
%  (2021), arXiv:2101.12345.
%
%\end{thebibliography}

%%%%%%%%%%%%%%%% ACKNOWLEDGEMENTS %%%%%%%%%%%%%%%

\section*{Acknowledgments}
\addcontentsline{toc}{section}{Acknowledgments}
We thank Professor Van Anh Ho and Quang Ngoc Pham for letting us test the foundation policy on the soft quadrotor. 
\paragraph*{Funding:}
This work was supported in part by the National Science Foundation (NSF) CAREER program under Grant 2145277, and in part by the Defense Advanced Research Projects Agency (DARPA) Young Faculty Award under Grant D22AP00156-00.
\paragraph*{Author Contributions:}
J.E. formulated the main ideas, implemented them, conducted the experiments and wrote the paper. 
D.A. and G.L. provided supervision and guided the direction during all phases of the project and helped write the paper.
\paragraph*{Competing Interests:}
The authors declare that they have no competing interests. 
\paragraph*{Data and Materials Availability:}
The code and dataset of sampled quadrotor dynamics as well as the trained teacher checkpoints are available from \url{https://doi.org/10.5281/zenodo.17096679} \cite{eschmann_2025_17096679} and the Git repository \url{https://github.com/rl-tools/raptor} \cite{git_repo}. Please also refer to the project page at \url{https://raptor.rl.tools} \cite{project_page} where an interactive simulation can be used to test the RAPTOR policy. The full-length, high-quality video can also be accessed at \url{https://youtu.be/hVzdWRFTX3k} \cite{video}.

%%%%%%%%%%%%%%%% SUPPLEMENT LIST %%%%%%%%%%%%%%%

% List the contents of your Supplementary Materials, including the numbers of any
% supplementary figures, tables, external data files etc. and any references that are
% cited only in the supplement. In this example, refs. 7-8 are cited only in the supplement.
% Fill out your numbers accordingly and delete any lines that aren't applicable.
\subsection*{Supplementary Materials}
\addcontentsline{toc}{subsection}{Supplementary Materials}
% Materials and Methods\\
% Supplementary Text\\
Figure S1\\
%Tables S1 to S4\\
% References \textit{(7-\arabic{enumiv})}\\ % automatically fills out the last reference number
% (filling out the other numbers automatically is possible but fiddly and liable to break)
Movie S1 to S8\\
Data S1

%%%%%%%%%%%%%%%% END OF MAIN TEXT %%%%%%%%%%%%%%%

\newpage

%%%%%%%%%%%%%%%% START OF SUPPLEMENT %%%%%%%%%%%%%%%

% Figures, tables, equations and pages in the supplement are numbered S1, S2 etc.
\renewcommand{\thefigure}{S\arabic{figure}}
\renewcommand{\thetable}{S\arabic{table}}
\renewcommand{\theequation}{S\arabic{equation}}
\renewcommand{\thepage}{S\arabic{page}}
\setcounter{figure}{0}
\setcounter{table}{0}
\setcounter{equation}{0}
\setcounter{page}{1} % not 0 as \newpage already started a supplementary page
% References continue the numbering from the main text.

%%%%%%%%%%%%%%%% SUPPLEMENT TITLE PAGE %%%%%%%%%%%%%%%

\begin{center}
\section*{Supplementary Materials for \vspace{2.1cm} \\ \scititle}
\addcontentsline{toc}{section}{Supplementary Materials}

% Author list for the supplement
% Indicate the corresponding authors, but do NOT include institutions here
% It would be nice if the template auto-generated this, but doing so is complicated...
Jonas Eschmann$^{\ast}$,
Dario Albani,
Giuseppe Loianno\\ % we're not in a \author{} environment this time, so use \\ for a new line
\small$^\ast$Corresponding author. Email: jonas.eschmann@berkeley.edu\\
\end{center}

% Fill out the numbers for each type of supplementary material,
% and delete any lines that aren't applicable.
% These are just example numbers that don't match the rest of this template.
\subsubsection*{This PDF file includes:}
Materials and Methods\\
Figure S1 \\
Captions for Movies S1 to S8\\
Captions for Data S1

\subsubsection*{Other Supplementary Materials for this manuscript:}
Movies S1 to S8\\
Data S1

\newpage

%%%%%%%%%%%%%%%% MATERIALS AND METHODS %%%%%%%%%%%%%%%

\subsection*{Materials and Methods}
\addcontentsline{toc}{subsection}{Materials and Methods}
\subsubsection*{Quadrotor Dynamics}
\addcontentsline{toc}{subsubsection}{Quadrotor Dynamics}

The L2F simulator \cite{eschmann2023learning} uses the following equations of motion:
\begin{align*}
\dot{\mathbf{p}} = & \; \mathbf{v} \\
\dot{\mathbf{q}} = & \; \mathbf q \circ \left[0 \;\; \boldsymbol{\omega}/ 2\right]^\top\\
\dot{\mathbf{v}} = & \; \frac{1}{m}\left(\mathbf{\mathbf{R}}(\mathbf{q}) \left(\sum_{i=0}^3 \mathbf{r}_{f_i} f_i\right) + \mathbf{f}_{\text{ext}}\right) + \mathbf{g} \\
f_i = & \; \sum_{j=0}^{2} c_{f_j} \left(\omega_{m_i}\right)^j \\
\dot{\boldsymbol{\omega}} = & \; \mathbf{J}^{-1}\left(\boldsymbol{\tau} +  \left(\mathbf{J} \boldsymbol{\omega}\right) \times \boldsymbol{\omega}\right) \\
\boldsymbol{\tau} = & \; \sum_{i=0}^3 \left(\mathbf{r}_{p_i} \times \mathbf{r}_{f_i}\right)f_i + \mathbf{r}_{\tau_i}c_m f_i  \\
\mathbf{r}_{p_i} = & \; \left[\pm \frac{l_\text{arm}}{\sqrt{2}}, \pm \frac{l_\text{arm}}{\sqrt{2}}, 0\right]^\top\\
\mathbf{r}_{\tau_i} = & \; \left[0, 0, \pm 1\right]^\top \\
\dot{\boldsymbol{\omega}}_{m_i} = & \; \begin{cases}
    T_{m\uparrow}^{-1} \left(\boldsymbol{\omega}_{{sp}_i} - \boldsymbol{\omega}_{m_i}\right), & \boldsymbol\omega_{m_i} \leq \boldsymbol\omega_{{sp}_i}  \\
    T_{m\downarrow}^{-1} \left(\boldsymbol{\omega}_{{sp}_i} - \boldsymbol{\omega}_{m_i}\right), & \boldsymbol\omega_{m_i} > \boldsymbol\omega_{{sp}_i}  \\
\end{cases} \\
\boldsymbol{\omega}_{sp} := & \; \mathbf{a} \\
\end{align*}
Where $\boldsymbol\omega$ is in the body frame. We only consider the case of planar quadrotors, where $\mathbf{r}_{f_i}=\left[0, 0, 1\right]^\top \; \forall i \in 0\ldots 3$. $\mathbf{r}_{p_i}$ and $\mathbf{r}_{\tau_i}$ are set up according to the Crazyflie motor layout conventions: (front-right, back-right, back-left, front-left) and (CCW, CW, CCW, CW), respectively. Note that this corresponds to the top-down view and the body frame is following the Front-Left-Up (FLU) conventions. CW corresponds to a torque direction of $\mathbf{r}_{\tau_i} = \left[0, 0, 1\right]$. When deploying on platforms with other motor-layout conventions, a re-mapping of the motor commands is applied.

\subsubsection*{Linear Velocity Feedback Delay Mitigation}
\addcontentsline{toc}{subsubsection}{Linear Velocity Feedback Delay Mitigation}

We counteract the linear velocity delay experienced by the non-EKF-based platforms by overlaying a simple accelerometer integral that is grounded through an exponential decay. This can also be interpreted as a convolution of the accelerometer data with an Infinite Impulse Response (IIR) filter. Before the convolution, the gravity is subtracted by applying the current orientation estimate to the accelerometer data.
\begin{align}
\mathbf{a}_g(t) &= \mathbf{R}^T(t) \mathbf{a}(t) - \mathbf{g} \label{eq:gravity_subtraction} \\
\mathbf{v}_{a}(t) &= \int_0^t \mathbf{a}_g(\tau) e^{-\frac{t-\tau}{T}} d\tau \label{eq:integral_decay}
\end{align}

We use the following discrete approximation to implement the filter:
\begin{align}
\alpha &= e^{-\frac{\Delta t}{T}} \\
\mathbf{v}_{a}^t &= \alpha \mathbf{v}_{a}^{t-1} + \mathbf{a}_g^{t} \Delta t
\end{align}
We find that this mitigation substantially reduces the z-axis oscillations on the non-EKF-based platforms.

\subsubsection*{Sampling Quadrotors}
\addcontentsline{toc}{subsubsection}{Sampling Quadrotors}
\label{sup:sampling-quadrotors}
We establish a physically plausible generative distribution over quadrotors through the following relations:
\begin{align}
r_{\mathrm{t2w}} & \sim \mathrm{Uniform}(1.5, 5) \\
m_{\mathrm{min}} & = 0.02, \quad m_{\mathrm{max}}= 5 \\
m & = s^3, \quad s \sim \mathrm{Uniform}(\sqrt[3]{m_{\mathrm{min}}}, \sqrt[3]{m_{\mathrm{max}}}) \\
f(\omega_{m_i}) & = c_{f_0} + c_{f_1} \omega_{m_i} + c_{f_2} \omega_{m_i}^2 \\
\quad \omega_{m_i} & \in [0, 1], \; \sum_i C_{f_i} = 1, \; 
C_{f_0} = 0.032,\; 
C_{f_1} = 0.131,\; 
C_{f_2} = 0.837\\ 
T & = r_{\mathrm{t2w}} \cdot 9.81 \cdot m  \\
c_{f_i} & = C_{f_i}  \frac{T}{4}\\
\end{align}

The idea is to first sample the thrust-to-weight ratio $r_{\mathrm{t2w}}$ and mass $m$ to create the thrust curve. The thrust curve consists of the coefficients $C_{f_0}, C_{f_1}, C_{f_2}$ for the constant, linear and quadratic term respectively. To relate the thrust-to-weight ratio to the thrust curve we require the mass. To sample the mass we need to consider that it grows cubically with the size (arm length, assuming constant density). If we were to uniformly sample the mass, we would disproportionately bias the distribution towards large quadrotors. Instead, we pose that quadrotors are rather uniformly distributed in size so we uniformly sample a scale $s$ that we map back cubically into the desired mass range $m \in [m_{\mathrm{min}}, m_{\mathrm{max}}]$  to get a realistic distribution over masses. Note that w.l.o.g., we normalize the motor effort setpoints $\omega_{m_i}$ and the baseline thrust curve coefficients $C_{f_i}$. This allows us to simply scale the baseline thrust curve shape (taken from the Crazyflie \cite{eschmann2024datadrivenidentificationquadrotorssubject}) to reflect the sampled thrust-to-weight ratio. Note: The baseline thrust curve coefficients $C_{f_i}$ are capitalized while the coefficients $c_{f_i}$ of the sampled thrust curve $f$ are lower-case.

Due to different design considerations, the mass-size ratio $\frac{\sqrt[3]{m}}{l_{\mathrm{arm}}}$ is not always constant so we establish the mass-size ratio of the Crazyflie $R_{ms}$ as the base value and sample variations around that. 

\begin{align}
R_{ms} & = \frac{\sqrt[3]{m_{\text{crazyflie}}}}{l_{\mathrm{arm},\mathrm{crazyflie}}} \approx 7.90 \\
u & \sim \mathcal{N}(-0.1, 0.1) \\
s_{ms} & = \begin{cases}
  \frac{1}{1-u} & \text{if } u < 0 \\
  1+u & \text{if } u \geq 0
\end{cases} \\
r_{ms} & = \frac{\sqrt[3]{m}}{l_{\mathrm{arm}}} \overset{!}{=} s_{ms}R_{ms} \\
l_{\mathrm{arm}} & = \frac{\sqrt[3]{m}}{s_{ms}R_{ms}} \\
\end{align}

We intended to use a reciprocal deviation $s_\mathrm{ms}$ of $\pm 10\%$ but accidentally used a normal instead of a uniform distribution. In a post-hoc analysis we find that the resulting distribution has a mean and standard deviation of  $\approx 7.24 \pm 0.66$. This incidentally fits the empirical distribution of mass-size deviations of the quadrotors in Figure \ref{fig:quadrotor-table} with a mean and standard deviation of $\approx 7.66 \pm 1.93$ better than the intended distribution of just $10\%$ uniform reciprocal deviations. For future works we advise to investigate directly sampling from a normal distribution with a wider standard deviation that is closer to the empirical standard deviation of the actual quadrotors. 

The sampled mass-size ratio allows us to compute the arm length $l_{\mathrm{arm}}$ which, in combination with the assumption that the quadrotor is in a symmetric X configuration, defines the geometry. 

Finally, we need to sample the inertia based on the previously sampled quantities:
\begin{align}
r_{\mathrm{t2i}} & \sim \mathrm{Uniform}(40, 1200) \\
\tau & = T \cdot \sqrt{2} \cdot l_{\mathrm{arm}}  \quad \\ 
J_{xx} & = J_{yy} = \frac{\tau}{r_{\mathrm{t2i}}} \\
J_{zz} & = \frac{J_{xx} + J_{yy}}{2} \cdot 1.832 \\
\end{align}

For sampling the inertia, we introduce another ratio, the torque-to-inertia ratio $r_{\mathrm{t2i}}$. We collect quadrotor dynamics parameters from the literature and find a realistic randomization range to be between $40$ and $1200$. After sampling the torque-to-inertia ratio, we can use the previously sampled thrust-to-weight, mass and size (arm length) parameters to calculate the x and y inertia. To get the z inertia, we apply the rule of \cite{eschmann2024datadrivenidentificationquadrotorssubject}.

Furthermore, we can independently sample the moment constant as well as the motor delays because they do not correlate strongly with any of the other variables in our experience:

\begin{align}
c_{m} & \sim \mathrm{Uniform}(0.005, 0.05) \\
T_{m \uparrow} & \sim \mathrm{Uniform}(0.03, 0.1) \\
T_{m \downarrow} & \sim \mathrm{Uniform}(0.03, 0.3)
\end{align}

Finally, we sample the standard deviation used for sampling the random disturbance force $\mathbf{f}_{\text{ext}}$ as $10\%$ of the surplus thrust $r_{\text{t2w}}-1$:

\begin{align}
\sigma_{\mathbf{f}_{\text{ext}}} & \sim \mathrm{Uniform}(0, (r_{\text{t2w}}-1)\cdot 0.1)
\end{align}

The idea behind choosing $10\%$ is that by applying the $3 \sigma$-rule, the sampled random disturbance will be confined at $30\%$ of the surplus thrust with a $\sim99.7\%$ probability. This makes it exceedingly unlikely that we sample forces that are too strong to be compensated by the control policy.

\subsubsection*{Sampling Trajectories}
\addcontentsline{toc}{subsubsection}{Sampling Trajectories}
\label{sup:sampling-trajectories}
During pre-training, the policies are Markovian and only consider the current observation (plus previous actions). This means that (in the absence of aerial drag-forces from linear velocity) trajectory tracking and position control appear identical to the policy because we can just feed the position and linear velocity error w.r.t. the trajectory as observations and achieve good trajectory tracking \cite{eschmann2023learning}. In the case of a stateful, non-Markovian policy as with the recurrent foundation policy, this is not the case anymore. If we add the dynamics of the trajectory itself to the dynamics of the quadrotor through the error-state observation, the trajectory of error-state observations does not appear like a quadrotor anymore. 

A simple example would be a reference trajectory with a jump in linear velocity. From the perspective of a stateful policy that has only been trained on position control (no trajectory dynamics in the observation space) this abrupt change in linear velocity appears like a quadrotor with an unrealistically high thrust-to-weight ratio, unrealistically fast angular response and/or even a world with much larger than $1$ G gravity. The latter happens if, for example, the velocity step is downwards in z and the quadrotor is not upside down or if the action inputs have been low/zero.

To counter-act this, we train the expert and student policies using a simple probabilistic mixture model of trajectories. With $50\%$ probability, the task is just tracking the null-trajectory (position control, going back to the origin from any initial state). In the other $50\%$ the task is to track randomly sampled trajectories. Note that position and linear velocity are just offset by the trajectory ($\mathbf{p} := \mathbf{p} - \mathbf{p}_{\text{target}}$, $\mathbf{v} := \mathbf{v} - \mathbf{v}_{\text{target}}$) in the observations as well as in the reward function. We neither change the structure nor adjust the parameters. 

We would like to cover a wide variety of possible reference trajectories. Trajectories are vector-valued functions of time, so we need to design a broad distribution over functions. Here we take inspiration from the Gaussian Process (GP) community, which has been concerned with designing prior distributions over functions since its inception. Additionally, we require the sampling of reference trajectories to not slow down the simulation too much. Hence, we choose to sample the reference trajectories from a second-order Langevin process. A second-order Langevin process corresponds to a GP with a certain structure in the kernel (depending on the parameters of the Langevin process) \cite{langevin-gp-correspondence,sarkka2019applied} but we can easily sample it incrementally while simulating the quadrotor dynamics. 

Based on the results described in the Trajectory Tracking subsection, we find that this approach works well for moderately agile trajectory tracking and even generalizes from second-order Langevin random walks to cyclical Lissajous trajectories. But we acknowledge that, although the second-order Langevin process covers the space of smooth functions relatively well, many trajectories with real-world use-cases such as step-functions/responses are not covered or exponentially unlikely (such as cyclical trajectories).

\subsubsection*{Meta-Imitation Learning Objective}
\addcontentsline{toc}{subsubsection}{Meta-Imitation Learning Objective}

We do not perform variational inference on a case-by-case basis but instead in an amortized fashion, where the inference itself is conducted by the recurrent neural network policy at test time. This amortization makes the inference very compute efficient and allows us to deploy the foundation policy onto even the tiniest microcontrollers while meeting real-time constraints at high frequencies.

For tractability of the KL divergence we assume the action distributions are parameterized Gaussians: 
\begin{align}
p(\mathbf{a}_t^* \mid \mathbf{s}_t, \boldsymbol\Xi) \approx \; &  \mathcal{N}(\mathbf{a}_t^*; \boldsymbol\pi^*(\mathbf{s}_t, \boldsymbol\Xi), \mathbf{I}) \\
\boldsymbol\pi^*(\mathbf{s}_t, \boldsymbol\Xi) := \; & \boldsymbol\pi^*_{\boldsymbol\Xi}(\mathbf{s}_t) \\
p(\mathbf{a}_t^\pi \mid \mathbf{o}_0, \ldots, \mathbf{o}_t, \mathbf{a}_0, \ldots, \mathbf{a}_{t-1}) \approx \;  & \mathcal{N}(\mathbf{a}_t^\pi; \boldsymbol\pi(\mathbf{o}_0, \ldots, \\
& \mathbf{o}_t, \mathbf{a}_0, \ldots, \mathbf{a}_{t-1}), \mathbf{I}) \\
= \; & \mathcal{N}(\mathbf{a}_t^\pi; \boldsymbol\pi(\mathbf{o}_{0:t}, \mathbf{a}_{0:t-1}), \mathbf{I})
\end{align}
Where $\boldsymbol\pi^*_{\boldsymbol\Xi}$ is one of the $1000$ teacher policies trained for the particular quadrotor/set of dynamics parameters $\boldsymbol\Xi$ and $\boldsymbol\pi$ is the student foundation policy.

We can see that the input shape of the foundation policy $\boldsymbol\pi$ is dependent on the number of previous steps in the episode. Due to the sequential nature of the inputs, we choose a recurrent neural network architecture for three main reasons: 
\begin{enumerate}
\item \textbf{Computational Efficiency}: We require computational efficiency for direct deployment on compute-constrained microcontrollers. Recurrent Neural Networks (RNNs) are $\mathcal{O}(1)$ in the history length $N$, each incremental step only requires a fixed amount of compute at inference time. In contrast, Convolutional Neural Networks (CNN) and attention are $\mathcal{O}(\log(N))$ and $\mathcal{O}(N)$ for each step, respectively. 
\item \textbf{Context Window Extrapolation}: At inference time we would like the policy to fly for longer time than the context window used during training. During training the context is usually limited to not slow down the process too much. CNNs, by design, have a limited context window that cannot be extended. Similarly, attention, being a set-to-set mapping, requires position embeddings which complicate context window extrapolation. 
\item \textbf{Successful Use in Related Works}: E.g. \cite{openai_solving_2019} and \cite{berner2019dota} have used recurrent policies successfully.
\end{enumerate}

We find that a surprisingly small (in terms of number of parameters and compute requirements) three-layer recurrent neural network is sufficient to express the desired behavior described in the Introduction section. Figure \ref{fig:probabilistic-model-arch-concept}B shows the architecture which contains a dense input layer, Gated Recurrent Unit (GRU) layer and a dense output layer. The initialization of the recurrent state (value at step $0$) is all zeros and we feed back the previous output of the policy as an input. Due to the small hidden dimensions the foundation policy only has $2084$ parameters:
\begin{align}
P = & \;  P_\text{input} + P_\text{GRU} + P_\text{output} \\ 
P_\text{input} = & \; 22 \cdot 16 + 16 = 368 \\
P_\text{GRU} = & \; 16 \cdot 16 \cdot 3 \cdot 2 + 16 \cdot 3 \cdot 2 + 16 = 1648 \\
P_\text{output} = & \; 16 \cdot 4 + 4 = 68 \\
P = & \; 2084
\end{align}

We train the foundation policy using Meta-Imitation Learning where it acts as the student. During Meta-Imitation Learning, we want to adjust the student's weights to minimize the KL divergence (also known as relative entropy) between the predicted optimal action distribution by the student and the optimal action distribution (approximated by the particular teacher for the current system dynamics):

\begin{align}
D_{\mathrm{KL}}\!\bigl(\mathbf{a}_t^* \,\|\, \mathbf{a}_t^\pi\bigr) = & \; \underset{\mathbf{a}_t^* \sim p(\mathbf{a}_t^* \mid \mathbf{s}_t, \boldsymbol\Xi)}{\mathbb{E}} \left[\log\frac{p(\mathbf{a}_t^* \mid \mathbf{s}_t, \boldsymbol\Xi)}{p(\mathbf{a}_t^\pi = \mathbf{a}_t^* \mid \mathbf{o}_{0:t}, \mathbf{a}_{0:t-1})}\right] \label{eq:kl}\\
\left[\ldots\right]= & \; \log p(\mathbf{a}_t^* \mid \mathbf{s}_t, \boldsymbol\Xi) \\ 
& \; - \log p(\mathbf{a}_t^\pi = \mathbf{a}_t^* \mid \mathbf{o}_{0:t}, \mathbf{a}_{0:t-1}) \\
\end{align}
Hence, we want to find the log probabilities of the action distributions that we approximated as Gaussians before:  
\begin{align}
\mathcal{N}(\mathbf{x}; \boldsymbol\mu, \boldsymbol\Sigma) = & \; (2\pi)^{-k/2}\det(\boldsymbol\Sigma)^{-1/2} \\ 
& \; \cdot \exp\!\left( -\frac12 (\mathbf{x}-\boldsymbol\mu)^{\mathsf T}\boldsymbol\Sigma^{-1}(\mathbf{x}-\boldsymbol\mu) \right) \\
\mathcal{N}(\mathbf{x}; \boldsymbol\mu, \mathbf{I}) = & \; C \cdot \exp\!\left( -\frac12 (\mathbf{x}-\boldsymbol\mu)^{\mathsf T}(\mathbf{x}-\boldsymbol\mu) \right) \\
 = & \; C \cdot \exp\!\left( -\frac12 \| \mathbf{x}-\boldsymbol\mu \|_2^2 \right) \label{eqn:kl-mse-squared-form}
\end{align}
Since we are aiming at quadrotors, both, the numerator and denominator multivariate Gaussian are 4-dimensional and the constants cancel. 
\begin{align}
\left[\ldots\right]= & \; \log p(\mathbf{a}_t^* \mid \mathbf{s}_t, \boldsymbol\Xi) \\ 
& \; - \log p(\mathbf{a}_t^\pi = \mathbf{a}_t^* \mid \mathbf{o}_{0:t}, \mathbf{a}_{0:t-1}) \\
\boldsymbol\mu_T := & \boldsymbol\pi^*(\mathbf{s}_t, \boldsymbol\Xi) \\
\boldsymbol\mu_S := & \boldsymbol\pi(\mathbf{o}_{0:t}, \mathbf{a}_{0:t-1}) \\
\left[\ldots\right] = & \; \log C - \frac{1}{2} \| \mathbf{a}^*_t-\boldsymbol\mu_T \|_2^2 \\ 
& \; - \log C + \frac{1}{2} \| \mathbf{a}^*_t-\boldsymbol\mu_S \|_2^2 \\ 
\end{align}
We can split the second squared norm:
\begin{align*}
& \| \mathbf{a}^*_t-\boldsymbol\mu_S \|_2^2 \\
= &\;   \| (\mathbf{a}^*_t-\boldsymbol\mu_T) + \left(\boldsymbol\mu_T - \boldsymbol\mu_S\right) \|_2^2 \\
= & \; \| (\mathbf{a}^*_t-\boldsymbol\mu_T)\|_2^2 + 2 (\mathbf{a}^*_t - \boldsymbol\mu_T)^{\mathsf T}(\boldsymbol\mu_T - \boldsymbol\mu_S) +  \| (\boldsymbol\mu_T-\boldsymbol\mu_S)\|_2^2 \\
\end{align*}
Plugging back into Eq. \ref{eqn:kl-mse-squared-form}:
\begin{align*}
\left[\ldots\right] = & \; (\mathbf{a}^*_t - \boldsymbol\mu_T)^{\mathsf T}(\boldsymbol\mu_T - \boldsymbol\mu_S) +  \frac12 \| (\boldsymbol\mu_T-\boldsymbol\mu_S)\|_2^2 
\end{align*}
We remember that we take the expectation over $\left[\ldots\right]$ in Eq. \ref{eq:kl} and that we assume that the teacher is an unbiased estimator for the optimal action ($\mathbb{E}\left[\mathbf{a}^*_t - \boldsymbol\mu_T\right] = 0$):
\begin{align}
D_{\mathrm{KL}}\!\bigl(\mathbf{a}_t^* \,\|\, \mathbf{a}_t^\pi\bigr) = & \; \frac12 \| (\boldsymbol\mu_T-\boldsymbol\mu_S)\|_2^2 \\
= & \; \frac12 \| (\boldsymbol\pi^*(\mathbf{s}_t, \boldsymbol\Xi)-\boldsymbol\pi(\mathbf{o}_{0:t}, \mathbf{a}_{0:t-1}))\|_2^2 \label{eq:kl-mse-loss-function}
\end{align}

Hence, under the mild assumption of unit standard deviations, we can conduct the Meta-Imitation Learning in a principled manner by minimizing the relative differential entropy between the (approximated) optimal action distribution and the predicted optimal action distribution by the student. In practice, we implement this by using Eq. \ref{eq:kl-mse-loss-function}, which is identical to the Mean-Squared Error (MSE), as the loss function and by optimizing the student policy's parameters $\boldsymbol\theta$ using gradient descent.

%%%%%%%%%%%%%%%% SUPPLEMENTARY TEXT %%%%%%%%%%%%%%%

% \subsection*{Supplementary Text}
% The Supplementary Text section can only be used to directly support statements made in the main text
% e.g. to present more detailed justifications of assumptions, investigate alternative scenarios,
% provide extended acknowledgements etc.
% Material in this section cannot claim results or conclusions that weren't mentioned in the main text.
% To refer to this section from the main text, just write (Supplementary Text).
% 
% \subsubsection*{Example supplement heading}

% The two main sections of the supplement can be split up using headings.

% If your supplement is very short you might need to uncomment the following line to avoid
% layout problems with the figures and tables.
%\newpage

%%%%%%%%%%%%%%%% SUPPLEMENTARY FIGURES %%%%%%%%%%%%%%%
\begin{figure}
        \centering
        \includegraphics[width=0.5\textwidth]{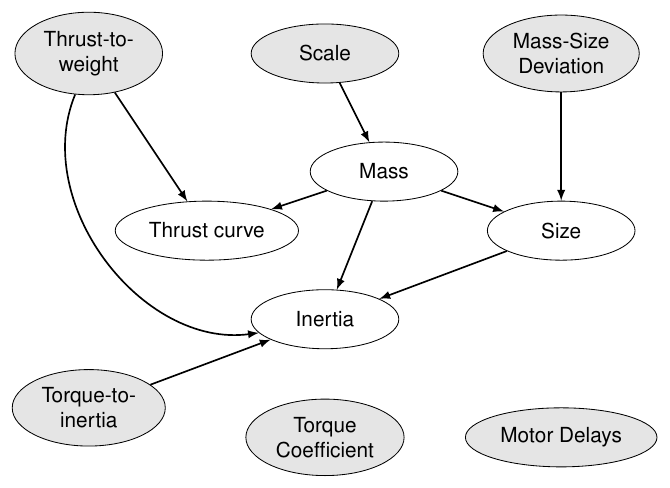}
        \caption{\textbf{Probabilistic Graphical Model for Ancestral Sampling of Quadrotors}}
        \label{fig:sampling-quadrotors-graphical-model}
\end{figure}

%%%%%%%%%%%%%%%% SUPPLEMENTARY TABLES %%%%%%%%%%%%%%%

\newpage
%%%%%%%%%%% CAPTIONS FOR OTHER SUPPLEMENTARY FILES %%%%%%%%%%

% \clearpage % Clear all remaining figures and tables then start a new page
% 
\paragraph{Caption for Movie S1.}
\textbf{Motivation and Introduction}
\paragraph{Caption for Movie S2.}
\textbf{Trajectory Tracking Experiments}
\paragraph{Caption for Movie S3.}
\textbf{Outdoor Experiments}
\paragraph{Caption for Movie S4.}
\textbf{Disturbance Experiments: Poking}
\paragraph{Caption for Movie S5.}
\textbf{Disturbance Experiments: Wind}
\paragraph{Caption for Movie S6.}
\textbf{Disturbance Experiments: Different Propellers}
\paragraph{Caption for Movie S7.}
\textbf{Agile Recovery / Rapid In-Context Learning Experiments}
\paragraph{Caption for Movie S8.}
\textbf{Yaw Response Experiments}

\paragraph{Caption for Data S1.}
\textbf{Data and Code}
This package \cite{eschmann_2025_17096679} contains all training and inference code as well as the complete pre-training data (including dynamics parameters and checkpoints of the $1000$ quadrotors) and the resulting foundation policy. Please refer to the readme.txt for instructions.

%%%%%%%%%%%%%%%% SUPPLEMENTARY REFERENCES %%%%%%%%%%%%%%%

% Do NOT include a reference list in the supplement.
% All references must be in a single list at the end of the main text.
% The copyeditors will ensure that the correct reference list appears with each version of the paper
% (print, HTML, PDF, mobile app, metadata for bibliographic databases etc.)

\end{document}